%% file: 21-BP-GNN-SSP-V11.tex
\documentclass{article}
\usepackage[preprint]{spconf}
\usepackage{amsmath}
\usepackage{amssymb}
\usepackage{graphicx}
\usepackage{color}
\usepackage{cite}
\usepackage{bm}
\usepackage{epsfig,psfrag}
\usepackage{tabularx}
\usepackage{acronym}
\usepackage[yyyymmdd,hhmmss]{datetime}
\usepackage{bbm}
\usepackage{mathrsfs} 
\usepackage{bardiag}
\usepackage{tabularx}
\usepackage{multirow}
\usepackage{colortbl,pgfplotstable}
\usepackage{LatexInclusion/notation}
\usepackage[level=3]{LatexInclusion/messages}  
\usepackage{subfig}
\usepackage{enumitem}
\usepackage{float}
\usepackage{epsfig,psfrag}

\input{LatexInclusion/defmetric.tex}

\ninept

\allowdisplaybreaks
\definecolor{BLUE}{rgb}{0,0,1}

\providecommand{\bl}{\textcolor{blue}}
\providecommand{\rd}{\textcolor{red}}

\providecommand{\ist}{\hspace*{.3mm}}

\providecommand{\rmv}{\hspace*{-.3mm}}

\providecommand{\nn}{\nonumber}

\acrodef{bp}[BP]{belief propagation}
\acrodef{nebp}[NEBP]{neural enhanced belief propagation}
\acrodef{cl}[CL]{cooperative localization}
\acrodef{gnn}[GNN]{graph neural network}
\acrodefplural{gnn}[GNNs]{Graph neural networks}
\acrodef{ls}[LS]{least square}
\acrodef{ml}[ML]{maximum likelihood}
\acrodef{mpnn}[MPNN]{message passing neural network}
\acrodef{2d}[2-D]{two dimensional}
\acrodef{mmse}[MMSE]{minimum mean square error}
\acrodef{mlp}[MLP]{multilayer perceptron}
\acrodef{nees}[NEES]{normalized estimation error squared}
\acrodef{pdf}[PDF]{probability density function}

\newcommand{\acr}[1]{\aclu{#1} (\acs{#1})}

\newcommand{\T}{\text{T}}
 
\name{Mingchao Liang and Florian Meyer \vspace{-2mm}}
\address{\normalsize University of California, San Diego, La Jolla, CA\\[-.5mm] \normalsize Email: \{m3liang, flmeyer\}@ucsd.edu \vspace{-3mm}}

\definecolor{BLUE}{rgb}{0,0,1}

\definecolor{myred}{rgb}{1,0.27,0}
\definecolor{mygreen}{rgb}{0.1, 0.55, 0.1}
\definecolor{myblue}{rgb}{0, 0, 1}

\newdateformat{monthyeardate}{\monthname[\THEMONTH] \THEDAY, \THEYEAR} 

\newcommand{\paperTitle}{\vspace{-3mm} Neural Enhanced Belief Propagation for Cooperative Localization\\
}

\newcommand{\BigPicture}{Location awareness \cite{BarRonKir:01,PatAshKypHerMosCor:05,BahWalLeo:C09,TarMupRauSloSveWym:J14,IhlFisMosWil:05,WymLieWin:09,WinMeyLiuDaiBarCon:J18,MeyKroWilLauHlaBraWin:J18} is an important aspect in a variety of applications including autonomous navigation, applied ocean sciences, and public safety.
}

\newcommand{\ClassicApproach}{Of particular interest are algorithmic solutions based on the framework of factor graphs and \ac{bp} \cite{IhlFisMosWil:05,WymLieWin:09,MeyKroWilLauHlaBraWin:J18,WinMeyLiuDaiBarCon:J18} due to their ability to provide accurate results in high-dimensional nonlinear Bayesian estimation problems. 
}

\newcommand{\SOADeficiency}{BP \cite{KscFreLoe:01,WeiFre:01, LiWu:19} is a message passing algorithm. It operates on the factor graph that represents the statistical model of an estimation problem. Given that the underlying factor graph is tree-structured, \ac{bp} is guaranteed to provide the exact marginal posterior distributions or ``beliefs'' needed for optimal estimation. However, in cases where the factor graph has cycles or the statistical model represented by the factor graph does not accurately model the true data generating process, \ac{bp} can only provide approximations of the marginal posterior distributions. In factor graphs with cycles, \ac{bp} is typically faced by a lack of convergence guarantees. \ac{bp} is also known to provide beliefs that are overconfident \cite{WeiFre:01}, i.e.,  the spread of the provided beliefs downplays the uncertainty of the estimates. This is particularly problematic in autonomous navigation applications where overconfidence can lead to catastrophic events \cite{BahWalLeo:C09}. }

\newcommand{\FundamentalQ}{We aim to improve the accuracy and reliability of BP-based localization and tracking algorithms by learning a refined model from data. }

\newcommand{\Advocacy}{A \ac{gnn} \cite{GorMonSca:05, ScaGorTsoHagMon:09} is a type of neural network that implements a message passing mechanism similar to \ac{bp}. It has been demonstrated that a learned \ac{gnn} can outperform loopy \ac{bp} for Bayesian estimation if sufficient data is available\cite{YooLiaXioZhaFetUrtZemPit:19}. Recently, \cite{SatWel:20} introduced \ac{nebp} which pairs a factor graph with a \ac{gnn}. The learned \ac{gnn} messages complement the corresponding \ac{bp} messages to correct errors introduced by cycles and model mismatch. The resulting method combines the benefits of model-based and data-driven inference. \ac{nebp} can provide satisfactory estimation results when little data is available and leverages the performance advantages of GNNs in the large data regime. So far \ac{nebp} has only been considered for estimation problems with discrete random variables. }

\newcommand{\ReasonSuccess}{}

\newcommand{\HLContribution}{In this paper, we extend \ac{nebp} to estimation problems with continuous random variables and apply it to the \ac{cl} problem \cite{IhlFisMosWil:05,WymLieWin:09,MeyHliHla:J14,WinMeyLiuDaiBarCon:J18}. In particular, we represent the beliefs and messages related to continuous random variables by random samples or ``particles'' and update their weights by combining the \ac{bp} message provided by the factor graph with the corresponding message provided by the \ac{gnn}. Compared to \ac{bp}-based \ac{cl}, the proposed \ac{nebp} method has an improved estimation accuracy and can avoid overconfident beliefs.
} 

\newcommand{\KeyContribution}{The main contributions of this paper are as \vspace{.5mm} follows.
    \begin{itemize}
        \item We extend \ac{nebp} to continuous random variables and apply it to the \ac{cl} \vspace{.5mm} problem.
        \item We demonstrate performance advantages compare to \ac{bp}-based \ac{cl} with a relative small amount of training data.
    \end{itemize}
    Our method preserves the advantages of \ac{bp}-based methods for \ac{cl} in wireless networks \cite{IhlFisMosWil:05,WymLieWin:09,WinMeyLiuDaiBarCon:J18}, i.e., it is fully distributed and requires little communication overhead, while its computational complexity only differs by a constant factor.}

\pgfplotsset{compat=1.14}

\begin{document}
\title{\paperTitle \vspace*{.5mm}}

\maketitle

\begin{abstract}
Location-aware networks will introduce innovative services and applications for modern convenience, applied ocean sciences, and public safety. In this paper, we establish a hybrid method for model-based and data-driven inference. We consider a \ac{cl} scenario where the mobile agents in a wireless network aim to localize themselves by performing pairwise observations with other agents and by exchanging location information. A traditional method for distributed \ac{cl} in large agent networks is \ac{bp} which is completely model-based and is known to suffer from providing inconsistent (overconfident) estimates. The proposed approach addresses these limitations by complementing  \ac{bp} with learned information provided by a \ac{gnn}. We demonstrate numerically that our method can improve estimation accuracy and avoid overconfident beliefs, while its computational complexity remains comparable to \ac{bp}. Notably, more consistent beliefs are obtained by not explicitly addressing overconfidence in the loss function used for training of the \ac{gnn}.
\end{abstract}



\acresetall
\section{Introduction}
\label{sec:introduction}

\BigPicture 
\ClassicApproach

\SOADeficiency
\FundamentalQ 

\Advocacy 
\ReasonSuccess 

\HLContribution 

\KeyContribution


\section{Review of Particle-based BP for CL}

We briefly review particle-based BP for CL which will be the basis for the development of the proposed \ac{nebp} method.
 
\subsection{System Model and Problem Formulation}

We consider a wireless network that consists of $I$ mobile agents with indexes $i \in \Set{I} \triangleq \{1,\dots,I\}$. The topology of the agent network is described by the sets of neighbors $\Set{N}_i \subseteq \Set{I} \backslash \{i\}$. In particular, agent $i$ is able to communicate and perform measurements with agents  $j  \rmv\in\rmv \Set{N}_i$. The state $\V{x}_{i, n} = [\V{p}_{i, n}^{\T} \ist \V{v}_{i, n}^{\T}]^{\T}$ of agent $i \in \Set{I}$ at time $n \in \{0, 1, \dots\}$ comprises the current position $\V{p}_{i, n} \in \mathbb{R}^{d}$  and other motion-related parameters $\V{v}_{i, n}$.  Agent motion is modeled by the mobility \vspace{-.5mm} model
\begin{equation}
    \V{x}_{i, n} = f(\V{x}_{i, n - 1}, \V{q}_{i, n}) \label{eq:transition}
    \vspace{0mm}
\end{equation}
where $\V{q}_{i, n}$ is the driving noise with known \ac{pdf} $p(\V{q}_{i, n})$ that is assumed to be statistically independent across $i$ and $n$. From the state transition function  \eqref{eq:transition} we can directly obtain the state-transition \ac{pdf} $p( \V{x}_{i, n} | \V{x}_{i, n - 1})$.

At time $n$, agent $i$ exchanges information with neighboring agents $j \in \Set{N}_i$ and performs a pairwise measurements that are modeled as 
\begin{equation}
    \V{z}_{j \to i, n} = h(\V{x}_{j, n}, \V{x}_{i, n}, \V{r}_{j \to i, n}) \label{eq:measurement}
    \vspace{2.5mm}
\end{equation}
where $\V{r}_{j \to i, n}$ is the measurement noise with known \ac{pdf} $p(\V{r}_{j \to i, n})$ that is assumed to be statistically independent across edges $(j,i)$, $i \in \Set{I}, j \in \Set{N}_i$ and time $n$. From the measurement model \eqref{eq:measurement}, we can directly obtain the likelihood function $p(\V{z}_{j \to i, n} | \V{x}_{j, n}, \V{x}_{i, n})$.

Let $\V{x}_{n} \rmv = \rmv [\V{x}_{i, n}]_{i \in \Set{I}}$ and $\V{z}_n \rmv = \rmv [\V{z}_{j \to i, n}]_{i \in \Set{I}, j \in \Set{N}_i}$ be the joint state and measurement vectors at time $n$. Furthermore, we introduce $\V{x}_{0 : n} = [\V{x}^{\T}_0 \ist \cdots \ist  \V{x}^{\T}_n]^{\T}$ and $\V{z}_{1 : n} = [\V{z}^{\T}_{1} \ist \cdots \ist \V{z}^{\T}_n]^{\T}\rmv\rmv$. The goal of \ac{cl} is to estimate the states of the agents $\V{x}_{i, n}$, $i \rmv\in\rmv \Set{I}$ from the joint measurement vector $\V{z}_{1 : n}$ by using, e.g., the \ac{mmse} estimator $\hat{\V{x}}_{i, n}^{\text{MMSE}} = \int \V{x}_{i, n} \ist\ist p(\V{x}_{i, n} | \V{z}_{1 : n}) \mathrm{d}\V{x}_{i, n}$. Estimation relies on the marginal posterior distributions $p(\V{x}_{i, n} | \V{z}_{1 : n}) = \int p(\V{x}_{0 : n} | \V{z}_{1 : n}) \V{x}_{\sim i}$. However, direct marginalization from $p(\V{x}_{0 : n} | \V{z}_{1 : n})$ is infeasible as its computation complexity grows exponentially with the number of time steps and the number of \vspace{-1.5mm} agents. 

\subsection{\ac{bp} for \ac{cl}}
\vspace{-.2mm}

\ac{bp} \cite{KscFreLoe:01} aims to calculate marginal posteriors $p(\V{x}_{i, n} | \V{z}_{1 : n})$ efficiently by passing messages on the edges of the factor graph that represents the joint \ac{pdf} of an estimation problem. In tree-structured graphs, the beliefs provided by \ac{bp} are guaranteed to converge to the true marginal posterior distributions $p(\V{x}_{i, n} | \V{z}_{1 : n})$. In graphs with cycles, (i) there are typically no convergence guarantees but \ac{bp} can nevertheless often provide an accurate approximation of $p(\V{x}_{i, n} | \V{z}_{1 : n})$ \cite{KscFreLoe:01,WeiFre:01}; and (ii) there are many possible orders in which messages are computed (also known as the message schedules), and different orders may lead to different beliefs.

\ac{bp} for \ac{cl} can provide accurate approximations of marginal distributions at a computational complexity that scales linearly with the number of time steps and the agent network size. In particular, by assuming that  at time $n=0$  the agent states $\V{x}_{i,0}$, $i \in \Set{I}$ are statistically independent and by using Bayes rule, the joint posterior distribution $p(\V{x}_{0 : n} | \V{z}_{1 : n})$ factorizes according to 
\vspace{-2mm}
\begin{align}
    p(\V{x}_{0 : n} | \V{z}_{1 : n}) &\propto \ist \prod^{I}_{i  = 1} \ist\ist p(\V{x}_{i, 0}) \rmv  \prod_{n^\prime = 1}^{n}  p(\V{x}_{i, n^\prime} | \V{x}_{i, n^\prime - 1})  \nonumber \\[.5mm]
    & \hspace{7.5mm} \times \prod_{j \in \Set{N}_i} p(\V{z}_{j \to i, n^\prime} | \V{x}_{j, n^\prime}, \V{x}_{i, n^\prime}) . \nn\\[-5.2mm]
    \nn 
\end{align} 

\begin{figure}
    \centering
    \subfloat[Factor Graph]{
    \psfrag{g1}[c][c][1]{\raisebox{-1.5mm}{\hspace{.4mm}$p_{1}$}}
    \psfrag{g2}[c][c][1]{\raisebox{-1.5mm}{\hspace{.4mm}$p_{2}$}}
    \psfrag{g3}[c][c][1]{\raisebox{-1.5mm}{\hspace{.4mm}$p_{3}$}}
    \psfrag{d1}[c][c][1]{\raisebox{-1.5mm}{\hspace{.5mm}$\V{x}_{1}$}}
    \psfrag{d2}[c][c][1]{\raisebox{-1.5mm}{\hspace{.5mm}$\V{x}_{2}$}}
    \psfrag{d3}[c][c][1]{\raisebox{-1.5mm}{\hspace{.5mm}$\V{x}_{3}$}}
    \psfrag{h12}[c][c][0.9]{$\hspace{.4mm} p_{1 \to 2}$}
    \psfrag{h21}[c][c][0.9]{$\hspace{.4mm} p_{2 \to 1}$}
    \psfrag{h13}[c][c][0.9]{$\hspace{.4mm} p_{1 \to 3}$}
    \psfrag{h31}[c][c][0.9]{$\hspace{.4mm} p_{3 \to 1}$}
    \psfrag{h23}[c][c][0.9]{$\hspace{.4mm} p_{2 \to 3}$}
    \psfrag{h32}[c][c][0.9]{$\hspace{.4mm} p_{3 \to 2}$}
    \psfrag{m31}[c][c][1.2]{\rd{$\mu_{3 \to 1}$}}
    \includegraphics[scale=0.75]{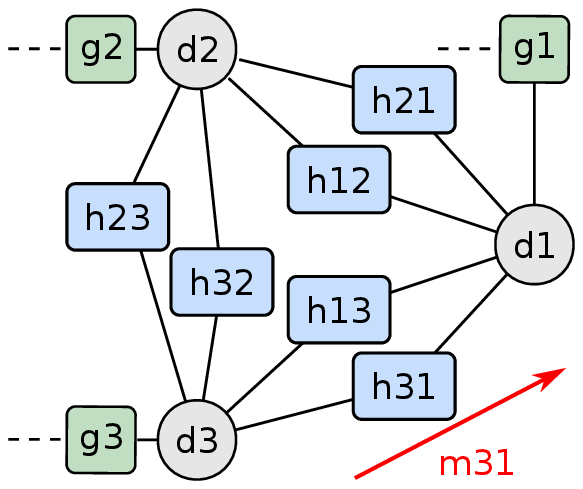}} 
    \hspace{4mm}
    \subfloat[\ac{gnn}]{
    \psfrag{d1}[c][c][1]{$\hspace{.5mm} \V{h}_{1}$}
    \psfrag{d2}[c][c][1]{$\hspace{.5mm} \V{h}_{2}$}
    \psfrag{d3}[c][c][1]{$\hspace{.5mm} \V{h}_{3}$}
    \psfrag{m31}[c][c][1.2]{\rd{$\V{m}_{3 \to 1}$}}
    \includegraphics[scale=0.75]{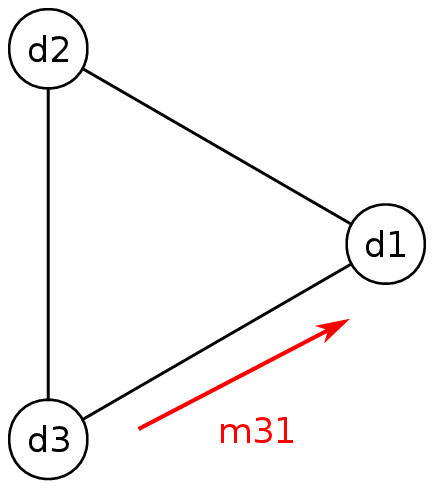}}
    \captionsetup{singlelinecheck = false, justification=justified}	
    \caption{Factor graph for \ac{cl} (a) and \vspace{.7mm} corresponding \acr{gnn} (b) for a\vspace{.7mm} single time slot $n$. The \ac{bp} and \ac{gnn} messages related to the messages $3 \rmv\to\rmv 1$ are also shown. The time index $n$ is omitted and the short notations $p_{j \to i} \overset{\Delta}{=} p(\V{z}_{j \to i, n} | \V{x}_{j, n}, \V{x}_{i, n})$, $p_{i} \overset{\Delta}{=} p(\V{x}_{i, n} | \V{x}_{i, n - 1})$, $\mu_{j \to i} = \mu_{j \to i, n}^{(t)}(\V{x}_{i, n})$, and $\V{m}_{j \to i} = \V{m}^{(t)}_{j \to i}$ are used.}
    \label{fig:fg_gnn}
    \vspace{-2.5mm}
\end{figure}

\vspace{-1.5mm} A  single time step of the corresponding cyclic factor graph is shown in Fig.~\ref{fig:fg_gnn}(a). 

This factor graph provides the basis for \ac{bp} for \ac{cl} where a specific message schedule makes it possible to perform message passing in real time and facilitates a distributed implementation. In particular, messages are sent only forward in time and $t \in \{1,\dots,T\}$ message passing iterations are performed at each time step $n$ individually. At each message passing iteration $t$, messages are passed only in one direction over every edge \cite{WymLieWin:09,WinMeyLiuDaiBarCon:J18}. The resulting \ac{bp} algorithm consists of prediction and update steps that are executed for each agent $i \in \Set{I}$ in \vspace{-.4mm} parallel.

\begin{itemize}[leftmargin=*]
    \item \textbf{Prediction Step}: Based on the belief $b_{i}^{(T)}\rmv(\V{x}_{i, n - 1})$ calculated at the previous time step $n - 1$ and the state transition \ac{pdf} $p(\V{x}_{i, n} | \V{x}_{i, n - 1})$, the ``prediction message'' $\mu_{i, \to n}(\V{x}_{i, n})$  is obtained \vspace{-2mm} as
    \begin{equation}
        \mu_{i, \to n}(\V{x}_{i, n}) \propto \int p(\V{x}_{i, n} | \V{x}_{i, n - 1}) \ist b^{(T)}_{i}(\V{x}_{i, n - 1}) \mathrm{d}\V{x}_{i, n - 1}. \label{eq:cl_precict}
    	\vspace{-.5mm}
    \end{equation}
    At time step $n\rmv=\rmv0$, message passing is initialized by\vspace{.3mm} setting $b_i^{(T)}\rmv(\V{x}_{i, 0})  \triangleq \vspace{.5mm} p(\V{x}_{i, 0})$.
    
    \item \textbf{Update Step}: At message passing iteration $t \in \{1,\dots,T\}$, beliefs $b_{j}^{(t - 1)}(\V{x}_{j, n})$ are received from neighboring agents $j \in \Set{N}_i$. Next, corresponding \vspace{.5mm} ``measurement messages'' $\mu_{j \to i, n}^{(t)}(\V{x}_{i, n})$ are calculated based on the likelihood function $p(\V{z}_{j \to i, n} | \V{x}_{j, n}, \V{x}_{i, n})$,\vspace{0mm} i.e.,
    \begin{equation}
        \mu_{j \to i, n}^{(t)}(\V{x}_{i, n}) \propto \int p(\V{z}_{j \to i, n} | \V{x}_{j, n}, \V{x}_{i, n}) \ist  b_{j}^{(t - 1)}(\V{x}_{j, n}) \mathrm{d}\V{x}_{j, n}. \label{eq:cl_update_msg}
    \end{equation}
    Finally, the belief at message passing iteration $t$ is obtained as 
    \begin{equation}
        b^{(t)}_{i}(\V{x}_{i, n}) \propto \mu_{i, \to n}(\V{x}_{i, n}) \prod_{j \in \Set{N}_i} \mu_{j \to i, n}^{(t)}(\V{x}_{i, n}). \label{eq:cl_update_belief}
    \end{equation} 
     \vspace{-0.5mm} At message passing iteration $t \rmv=\rmv 1$, beliefs are initialized as $b_{j}^{(0)}(\V{x}_{j, n}) = \mu_{i, \to n}(\V{x}_{j, n})$ (cf.~\eqref{eq:cl_update_msg}).
\end{itemize}

\subsection{Particle-Based Processing}

Often, the mobility and measurement models in \eqref{eq:transition} and \eqref{eq:measurement} are nonlinear and non-Gaussian and it is impossible to obtain a closed-form solution for the message passing and belief calculation equations in  \eqref{eq:cl_precict}--\eqref{eq:cl_update_belief}. This problem can be addressed by representing messages and beliefs by $K$ weighted random samples or ``particles'' and approximating \eqref{eq:cl_precict}--\eqref{eq:cl_update_belief} by means of Monte Carlo techniques \cite{MeyHliWymRieHla:16,MeyWymFroHla:J15}. The solution of the resulting particle-based processing can be arbitrary close to the corresponding true solution of  \eqref{eq:cl_precict}--\eqref{eq:cl_update_belief} by choosing $K$ sufficiently \vspace{-1mm} large.

\begin{itemize}[leftmargin=*]
    \item \textbf{Prediction Step}: 
Let $\big\{\V{x}_{i, n - 1}^{(k)}, w_{i, n - 1}^{(T, k)}\big\}_{k = 1}^K$ be a particle representation of $b_{i}^{(T)}(\V{x}_{i, n - 1})$. Then, a particle representation $\{\V{x}_{i, n}^{(k)}, w_{i, n}^{(k)}\}_{k = 1}^K$ of $\mu_{i, \to n}(\V{x}_{i, n})$ in \eqref{eq:cl_precict}, can be obtained by drawing, for each $k \rmv\in\rmv \{1,\dots,K\}$, one particle $\V{x}_{i, n}^{(k)}$ from $p(\V{x}_{i, n} | \V{x}_{i, n - 1}^{(k)})$, \vspace{-1.0mm} i.e.,
\vspace{-0.5mm}
\begin{equation}
    \V{x}_{i, n}^{(k)} \sim p(\V{x}_{i, n} | \V{x}_{i, n - 1}^{(k)}), \quad w_{i, n}^{(k)} = \vspace{-1mm} w_{i, n - 1}^{(T, k)}.  \nn
\end{equation}
At $n = 0$, we draw particles from \vspace{-1mm} the prior,  i.e. $ \V{x}_{i, 0}^{(k)} \sim p(\V{x}_{i, 0})$, and set corresponding weights according to $w_{i, 0}^{(T, k)} = \frac{1}{K}$.

\item \textbf{Update Step}:  Following the message multiplication scheme in \cite{MeyHliWymRieHla:16} a particle-based representation of $b^{(t)}_{i}(\V{x}_{i, n})$ in  \eqref{eq:cl_update_belief} is calculated for each message passing iteration $t \in \{1,\dots T\}$. In particular, particle-based messages
\vspace{.5mm}
\begin{equation}
\phi_{j \to i}^{(t,k)} \hspace{-.5mm} = \hspace{-.5mm} p(\V{z}_{j \to i, n} | \V{x}_{j, n}^{(k)}, \V{x}_{i, n}^{(k)}) \ist \tilde{w}_{j, n}^{(t-1, k)}
\label{eq:particleUpdate}
\vspace{.5mm}
\end{equation}
are\vspace{.5mm} computed for each neighbor $j \in \Set{N}_i$ and weight update is performed according \vspace{.3mm} to
\begin{equation}
    \tilde{w}_{i, n}^{(t, k)} = w_{i, n}^{(k)} \prod_{j \in \Set{N}_i} \phi_{j \to i}^{(t,k)}.  \label{eq:bp_particle_weight_update}
\vspace{-1mm}
\end{equation}
At message passing iteration $t\rmv=\rmv1$, \eqref{eq:particleUpdate} is initialized as  $\phi_{j \to i}^{(1,k)} \hspace{-.5mm} = \hspace{-.5mm} p(\V{z}_{j \to i, n} | \V{x}_{j, n}^{(k)}, \V{x}_{i, n}^{(k)}) \ist w_{j, n}^{(k)}$. Finally, particles and unnormalized \vspace{0mm} weights $\big\{\V{x}_{i, n}^{(k)}, \tilde{w}_{i, n}^{(t, k)}\big\}_{k = 1}^K$ are exchanged among neighboring \vspace{.2mm} agents. 
\end{itemize}

\vspace{-1mm} After the last iteration ($t\rmv = \rmv T$), the weights are normalized, i.e., $w_{i, n}^{(T, k)} = \tilde{w}_{i, n}^{(T, k)} / \sum_{k^\prime = 1}^K$ $\tilde{w}_{i, n}^{(T, k^\prime)} \rmv$, $k \in \{1,\dots,K\}$, and an\vspace{.4mm} approximation of the \ac{mmse} estimate can be obtained\vspace{.4mm} as $\hat{\V{x}}_{i, n} =$ $\sum_{k = 1}^{K} w_{i, n}^{(T, k)} \ist \V{x}_{i, n}^{(k)}$. The particle\vspace{.2mm} representation $\big\{\V{x}_{i, n}^{(k)}, w_{i, n}^{(T, k)}\big\}_{k = 1}^K$ is also needed for the prediction step at the next time step $n\rmv+\rmv1$.

For future reference, we introduce the weight vector  $\V{w}_{i} = [w_{i}^{(1)} \ist \cdots \ist w_{i}^{(K)}]^{\T}$ of agent $i$ after the prediction step and the joint weight vector $\V{w} \rmv=\rmv [\V{w}_{i} ]_{i \in \Set{I}}$. Similarly,\vspace{-0mm} we introduce the vectors of particle-based \ac{bp} messages $\V{\phi}_{j \to i}^{(t)} = \big[\phi_{j \to i}^{(t, 1) } \ist \cdots \ist \phi_{j \to i}^{(t, K)}\big]^{\T}$ and $\V{\phi}^{(t)} \rmv=\rmv \big[\V{\phi}_{j \to i}^{(t)} \big]_{i \in \Set{I}, j \in \Set{N}_i} \vspace{-1mm} $. 



\begin{figure*}[t!]
    \centering
\subfloat[\label{fig:realization_hybrid}]{\scalebox{0.3}{\input{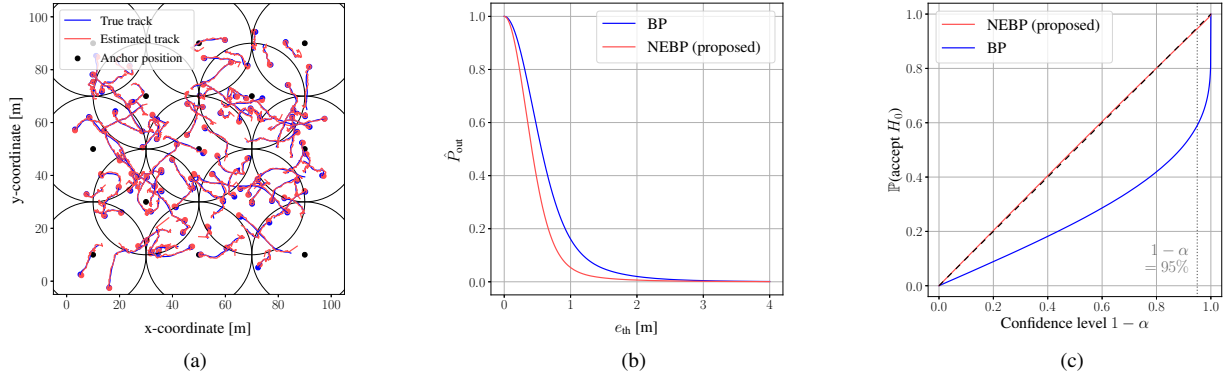}} \hspace{4mm}}
\hspace{7mm} 
\subfloat[\label{fig:outage}]{\scalebox{0.3}{\input{Figs/outage_r100.pgf}} \hspace{4mm}}
    \hspace{7mm}        
    \subfloat[\label{fig:confidence}]{\scalebox{0.3}{\input{Figs/confidence_percent_twosided_r100.pgf}} \hspace{4mm}}
  
    
    \captionsetup{singlelinecheck = false, justification=justified}
    \vspace{-1mm}
    \caption{Numerical evaluation of the proposed \ac{nebp} method in a \ac{cl} scenario with $13$ static anchors and $100$ mobile agents. (a) One realization of true and estimated mobile agent tracks with dots indicating the final positions. (b) Outage probability versus threshold $e_{\text{th}}$. (c) Probability that an estimate is consistent versus confidence levels. (The gray dotted-line indicates a confidence level of $95\%$.)}
\vspace{-2.5mm}
\end{figure*}

\section{NEBP for CL}
\vspace{-1mm}

In this section, we will review \acp{gnn} and present the proposed particle-based \ac{nebp} framework for \ac{cl}. In particular, at each time step $n$ we complement iterative particle-based \ac{bp} \eqref{eq:particleUpdate}--\eqref{eq:bp_particle_weight_update} by a GNN. Since we limit our discussion to a single time step, we will omit the time index $n$ in what follows. 

\subsection{Graph Neural Networks (GNNs)}
\acp{gnn} \cite{GorMonSca:05} extend neural networks to graph-structured data. We consider the \ac{mpnn} \cite{GilSchRilVinDah:17} which is a variant of \acp{gnn} that generalizes graph convolutional networks \cite{KipWel:17} and implements a message passing mechanism similar to \ac{bp}. A  \ac{mpnn} is defined on a graph $\Set{G} = (\Set{V}, \Set{E})$ where $\Set{E}$ induces the sets of neighbors $\Set{N}_i = \big\{j \in \Set{V} \big| (i,j) \in \Set{E} \big\}$. There is one neural network for each node and each edge in the graph. 
In many applications all node networks and all edge networks share sets of parameters, respectively.

Each node $i \in \Set{V}$ is associated with a vector $\V{h}_i$ called node embedding. At message passing iteration $t \in \{1,\dots,T\}$, the following operations are performed for each node $i \in \Set{V}$ in parallel. First, messages are exchanged with neighboring nodes $j \rmv\in\rmv \Set{N}_i$. In particular, the \ac{gnn} message sent from node $i \in \Set{V}$ to its neighbor $j \in \Set{N}_i$ is given \vspace{0mm} by
\begin{align}
    \V{m}_{i \to j}^{(t)} &= g_{\mathrm{e}} (\V{h}_{i}^{(t )}\rmv\rmv, \V{h}_{j}^{(t )}\rmv\rmv, \V{a}_{i \to j}) \label{eq:gnn_edge}\\[-3.5mm]
    \nn
\end{align}
where $g_{\mathrm{e}}(\cdot)$ is a neural network with trainable parameters \vspace{-.5mm} and  $\V{a}_{i \to j}$ is the edge attribute. Next, the node embedding $\V{h}^{(t)}_i$ is updated by incorporating the sum of received\vspace{.5mm} messages $\V{m}_{j \to i}^{(t)}$, $j \in \Set{N}_i$, i.e.,
\begin{align}
    \V{h}_{i}^{(t)} &= g_{\mathrm{n}} \Big(\V{h}_{i}^{(t - 1)}\rmv\rmv, \sum_{j \in \Set{N}_i} \V{m}_{j \to i}^{(t - 1)} \Big). \label{eq:gnn_node}
\end{align}
Here, $g_{\mathrm{n}} (\cdot)$ is again a neural network with trainable parameters. 

Since $g_{\mathrm{e}}(\cdot)$ and $g_{\mathrm{n}}(\cdot)$ share the same parameters across edges and nodes, respectively, they can be trained on small graphs even if then used in large scale inference problems. For future reference, we introduce the joint vector of node embeddings $\V{h} = [\V{h}_{i} ]_{i \in \Set{I}}$ and the joint vector of messages $\V{m} \rmv = \rmv [\V{m}_{j \to i}]_{i \in \Set{I}, j \in \Set{N}_i}$.




\subsection{Particle-based \ac{nebp}}
The main idea of particle-based \ac{nebp} is to use the particle representation $\big\{\V{x}_{i, n}^{(k)}, \tilde{w}_{i, n}^{(t, k)}\big\}_{k = 1}^K$ of a continuous state $\V{x}_{i, n}$ as if it would be the probability mass function (PMF) of a discrete random variable, i.e., the particles  $\V{x}_{i, n}^{(k)}$, $k \rmv\in\rmv\{1,\dots,K\} $ are the\vspace{-.5mm} possible outcomes and the particle weights  $w_{i, n}^{(k)}$ are the probabilities of these outcomes. 

The \ac{gnn} that is complementary to one time step of the \ac{cl} factor graph is shown in Fig.~\ref{fig:fg_gnn}(b).  Since the \ac{cl} factor graph only consists of pairwise interactions, we can use a simpler \ac{gnn} compared to the one originally proposed for \ac{nebp} \cite{SatWel:20}, i.e. a \ac{gnn} that only models variables nodes in the factor graph by a corresponding \ac{gnn} node. 
In what follows, we denote the \ac{nebp} messages and their joint vector by $\underline{\V{\phi}}^{(t)}$ and $\underline{\V{\phi}}_{j \rightarrow i}^{(t)}$, respectively.

\ac{nebp} for \ac{cl} consists of the following three \vspace{-.5mm} steps: 

\begin{enumerate}[leftmargin=*]

    \item First, classical \ac{bp} runs for one iteration,\vspace{.3mm} i.e.,
\begin{align}
    \V{\phi}^{(t)} &= \text{BP}\big( \underline{\V{\phi}}^{(t - 1)}\rmv\rmv, \V{w}\big). \label{eq:classical} \\[-6.5mm]
    \nn
\end{align}

Here $\text{BP}(\cdot)$ is the function that\vspace{-.6mm} takes \ac{nebp} messages $\underline{\V{\phi}}^{(t - 1)}$ and  $\V{w}$ as \vspace{.4mm} inputs and returns the \vspace{0mm} \ac{bp} messages $\V{\phi}^{(t)}$ by\vspace{0mm} first computing \eqref{eq:bp_particle_weight_update} \vspace{.6mm} (with $t$ replaced by $t\rmv-\rmv1$ and $\V{\phi}_{j \to i}^{(t,k)}$ replaced \vspace{-1mm} by $\underline{\V{\phi}}_{j \to i}^{(t-1,k)}$) followed by \eqref{eq:particleUpdate} for all edges $(j,i)$, $i \rmv\in\rmv \Set{I}$, $j \in \Set{N}_i$  in the network. At iteration $t \rmv=\rmv 1$, $\underline{\V{\phi}}^{(t - 1)}$ in \vspace{.3mm} \eqref{eq:classical} is replaced by\vspace{-1.5mm} the \vspace{.8mm} all-ones vector with dimension $K \ist \sum^{I}_{i = 1} |\Set{N}_i|$. 

\item Next, the output of the GNN is\vspace{0mm} computed,\vspace{.3mm} i.e.,
    \begin{align}
   \big[ \V{h}^{(t + 1) \T} \ist\ist\ist  \V{m}^{(t) \T} \big]^{\rmv\T} &= \text{GNN}\big(\V{h}^{(t)}\rmv\rmv, \V{\phi}^{(t)}\big) \nn \\[-5mm]
   \nn
\end{align}
    where $\text{GNN}(\cdot)$ is the function that calculates \ac{gnn} messages $\V{m}^{(t)}$ and updated node embeddings $\V{h}^{(t + 1)} \rmv\rmv$ from the current node embeddings $\V{h}^{(t)}$ and the classical \ac{bp} messages $\V{\phi}^{(t)}\rmv\rmv$ using \eqref{eq:gnn_edge}--\eqref{eq:gnn_node}. In \vspace{.5mm}  particular, the classical \vspace{0mm}\ac{bp} messages $\V{\phi}^{(t)}_{j \to i}$, $i \rmv\in\rmv \Set{I}$, $j \in \Set{N}_i$ are used as the edge attributes $\V{a}_{j \to i}$ in \eqref{eq:gnn_edge}. 

At $t\rmv=\rmv1$, $\V{h}^{(1)}$ is initialized by setting $\V{h}_{i}^{(1)} \rmv = \rmv \big [ \hat{\V{x}}_{i}^{\T} \ist\ist \text{vec}(\hat{\M{C}}_{i})^{\T} \big]^{\T}$ for all $i \in \Set{I}$, where $\hat{\V{x}}_{i} \rmv= \sum_{k = 1}^{K} w_{i}^{(k)} \V{x}_{i}^{(k)} $ is the sample mean, $\hat{\M{C}}_{i} \rmv= \sum_{k = 1}^{K} w_{i}^{(k)} (\V{x}_{i}^{(k)} - \hat{\V{x}}_{i}) (\V{x}_{i}^{(k)} - \hat{\V{x}}_{i})^{\T}$ is the sample \vspace{.3mm} covariance, and $\text{vec}(\cdot)$ creates a vector from the input matrix by taken elements columnwise.

\item Finally, for all edges $(j,i)$, $i \rmv\in\rmv \Set{I}$, $j \in \Set{N}_i$, \ac{gnn} messages $\V{m}^{(t)}_{j \to i}$ and \ac{bp} messages $\V{\phi}_{j \to i}^{(t)}$ are combined according to
\begin{align}
    \underline{\V{\phi}}_{j \to i}^{(t)} &= g_{\mathrm{s}}(\V{m}_{j \to i}^{(t)}) \ist \V{\phi}_{j \to i}^{(t)} + g_{\mathrm{v}}(\V{m}_{j \to i}^{(t)}). \nn
\end{align}

The functions $g_{\mathrm{s}}(\cdot)$  and $g_{\mathrm{v}}(\cdot)$  are neural networks with learnable parameters and output a positive scalar and a positive vector, respectively.
\end{enumerate}

\vspace{-1mm} After $T$ iterations, the particle representation $\{\V{x}_{i}^{(k)}\rmv, \underline{w}_{i}^{(T, k)}\}_{k = 1}^{K}$ of the \ac{nebp} belief $\underline{b}_{i}^{(T)}(\V{x}_{i})$ is obtained by calculating weights $\underline{w}_{i}^{(T, k)}$ based on \eqref{eq:bp_particle_weight_update} with \ac{bp} messages $\V{\phi}_{j \to i}^{(T,k)}$ replaced by \ac{nebp} messages $\underline{\V{\phi}}_{j \to i}^{(T,k)}$. The particle representation $\{\V{x}_{i}^{(k)}\rmv, \underline{w}_{i}^{(T, k)}\}_{k = 1}^{K}$ can then be used to calculate an approximate MMSE estimate $\underline{\hat{\V{x}}}_{i}$ and the corresponding approximate covariance matrix. Since we aim to reduce the MSE of the position estimate  $\underline{\hat{\V{p}}}_{i}$, the neural networks $g_{\mathrm{e}}(\cdot)$, $g_{\mathrm{n}}(\cdot)$, $g_{\mathrm{s}}(\cdot)$, and $g_{\mathrm{v}}(\cdot)$ are trained based on the loss function \vspace{0.2mm} $\sum^{I}_{i = 1} \rmv \|\underline{\hat{\V{p}}}_{i} - \V{p}_{i}\|^2\rmv$.



\section{Experiments}
\vspace{-1mm}

In this section, we compare the performance the proposed \ac{nebp} method with \ac{bp}. The neural networks $g_{\mathrm{e}}(\cdot), g_{\mathrm{s}}(\cdot), g_{\mathrm{v}}(\cdot)$ are \acp{mlp} with a single hidden layer and leaky rectified linear units (ReLUs) \cite{GooBenCou:16}, with exception that the output layers of $g_{\mathrm{s}}(\cdot)$ and $g_{\mathrm{v}}(\cdot)$ use sigmoid and ReLU activations, respectively. We set the number of message passing iterations to $T \rmv=\rmv 1$, in which case the node function $g_{\mathrm{n}}(\cdot)$ is not needed.  During training, the parameters of $g_{\mathrm{e}}(\cdot)$, $g_{\mathrm{s}}(\cdot)$, and $g_{\mathrm{v}}(\cdot)$ are updated through back-propagation. The dimension of node embeddings $\V{h}_i$ is $20$ and the dimension of \ac{gnn}\vspace{-.3mm}  messages $\V{m}_{j \to i}$ is \vspace{-2mm} $32$.

\subsection{Dataset and Training Procedure}
We consider agent networks in \ac{2d} space. The state of each agent at time $n$ is defined as $\V{x}_{i, n} = [\V{p}_{i, n}^{\T} \ist \V{v}_{i, n}^{\T}]^{\T} \in \mathbb{R}^4$ where $\V{p}_{i, n} \in \mathbb{R}^2$ and $\V{v}_{i, n} \in \mathbb{R}^2$ are the \ac{2d} position and velocity, respectively. We use a constant-velocity motion model with drag force and Gaussian driving noise with standard deviation $\sigma_a = 0.05$ (see \cite{VandeASte:15} for details).  Furthermore, we consider measurements of the distance $z_{j \to i, n} = \Vert \V{p}_{j, n} - \V{p}_{i, n} \Vert + r_{j \to i, n}$, where $r_{j \to i, n}$ is zero-mean Gaussian noise with standard deviation $\sigma_r = 1$. 

We consider $I \rmv=\rmv 25$ agents on the area of interest $[0, 60]\text{m} \times [0, 60] \text{m}$. There are five static anchors at perfectly known locations, i.e., their state transition model and prior distribution are given by $p(\V{x}_{i,n} | \V{x}_{i,n-1}) \rmv=\rmv \delta(\V{x}_{i,n} - \V{x}_{i,n-1})$ and $p\big(\V{x}_{i,0}) \rmv= \delta(\V{x}_{i,0}$ $- [\overline{\V{p}}_{i}^\T \ist\ist\ist 0 \ist\ist\ist 0 ]^{\T} \big)$ where $ \overline{\V{p}}_{i}$ is the true anchor position. In each realization, the mobile agents are uniformly placed over the area $[15, 45]\text{m} \times [15, 45] \text{m}$ and their velocity is randomly drawn from $\Set{N}(\V{0}, \sigma_p^2\M{I}_2)$ with $\sigma_p = 0.1$. For each agent, a track that consists of 50 time steps is generated and range measurements are obtained by assuming a connectivity of $20$m, i.e. $j \rmv\in\rmv \Set{N}_i$ if and only if $\Vert \V{p}_{j, n} - \V{p}_{i, n} \Vert \le 20\text{m}$. For inference, the initial prior distribution is $p(\V{x}_{i, 0}) = \Set{N}(\V{\mu}_{i, 0}, \M{\Sigma}_{i, 0})$. Here $\M{\Sigma}_{i, 0} = \text{diag}\{10, 10, 0.01, 0.01\}$ and $\V{\mu}_{i, 0}$ is randomly drawn from $\Set{N}(\V{x}_{i, 0}, \M{\Sigma}_{i, 0})$, where $\V{x}_{i, 0}$ is the true initial state of agent $i$. 

For training, an Adam optimizer \cite{KinBa:14} with learning rate $10^{-4}$ and batch size of $2$ is used. Furthermore, $100$ realizations of agent tracks and $10$ passes of the entire training dataset are considered. A larger agent network is employed for performance evaluation to show the generalization ability of our \ac{nebp} method. In particular, we consider a network that consists of  $13$ static anchors and $100$ mobile agents, i.e., $I = 100$ in the area $[0, 100]\text{m} \times [0, 100] \text{m}$ (see \cite{WymLieWin:09}) and generate another $400$ realizations of agents tracks. The agents are uniformly placed over $[10, 90]\text{m} \times [10, 90] \text{m}$ at time $n = 0$. All the other parameters are set as during  training. Anchor positions and a realization of mobile agent tracks are shown in \vspace{-3mm} Fig~\ref{fig:realization_hybrid}.

\subsection{Performance Evaluation}
\vspace{-.5mm}

To evaluate the performance of different localization algorithms, we use the outage probability $P_{\text{out}} = \mathbb{P} \big(\Vert \hat{\V{p}}_{i, n} - \V{p}_{i, n} \Vert > e_{\text{th}} \big)$, where $\V{p}_{i, n}$ is the true position, $\hat{\V{p}}_{i, n}$ is the estimate, and $e_{\text{th}} > 0$ is the error threshold. Fig. \ref{fig:outage} shows the outage probability versus threshold $e_{\text{th}}$. It can be seen that our proposed \ac{nebp} algorithm yields significantly reduced outage probability compared to \ac{bp}. Fig. \ref{fig:realization_hybrid} shows one realization of true and estimated agent tracks.

To assess the consistency of \ac{bp} and \ac{nebp} estimates, we conduct two-sided chi-square tests. We \vspace{-.4mm} first calculate the \ac{nees}  \cite{BarRonKir:01} \vspace{-.55mm}, $e_{i, n} = (\hat{\V{p}}_{i, n} - \V{p}_{i, n})^\T \hat{\M{\Sigma}}_{\V{p}, i, n}^{-1} (\hat{\V{p}}_{i, n} - \V{p}_{i, n})$,  where \vspace{.3mm} $\hat{\M{\Sigma}}_{\V{p}, i, n}^{-1} \in \mathbb{R}^{2 \times 2}$ is the estimated position covariance. Assuming that the true posterior distribution is Gaussian, the \ac{nees} follows a chi-square distribution with degree of freedom $2$. Let $H_0$ be the hypothesis that the belief of agent $i$ is\vspace{-.7mm} consistent, i.e., $\hat{\V{p}}_{i, n}$ and $\hat{\M{\Sigma}}_{\V{p}, i, n}$ are the true mean and covariance matrix. $H_0$ is accepted if $e_{i, n} \in [r_1, r_2]$, where $r_1, r_2$ is determined such that $\mathbb{P} ( e_{i, n} \le r_1 | H_0 ) = \mathbb{P} ( e_{i, n} \ge r_2 | H_0 ) = \frac{\alpha}{2}$ and $1 - \alpha$ is the confidence level. Ideally, as indicated by the black dashed line in Fig. \ref{fig:confidence}, at confidence level $1 - \alpha$, the probability that $H_0$ is accepted should be $1 - \alpha$. However, the \ac{bp} solution has $\mathbb{P} (\text{accept } H_0)$ significantly smaller than $1 - \alpha$. At $95\%$ confidence level, $40\%$ of the \ac{nees} values $e_{i,n}$ fall outside the confidence interval, indicating that \ac{bp} provides inconsistent estimates. On the other hand, the proposed \ac{nebp} method is close to the ideal line, where only $5\%$ of \ac{nees} values $e_{i,n}$ fall outside the $95\%$ confidence interval \cite{BarRonKir:01}. It can thus be concluded that \ac{nebp} significantly improves the consistency of estimates. Notably, more consistent estimates are obtained by not explicitly addressing overconfidence in the loss function used for training of the\vspace{-4mm} \ac{gnn}.

\section{Conclusion}
\vspace{-1mm}

In this paper, we propose a particle-based \ac{nebp} method for \ac{cl} that combines the benefits of model-based and data-driven inference. The proposed approach complements \ac{bp} with learned information provided by a \ac{gnn}.  Simulation results show that \ac{nebp} outperforms traditional \ac{bp} in terms of localization error and consistency of estimates as well as generalizes to larger agent networks.

\pagebreak

\renewcommand{\baselinestretch}{1}
\selectfont
\bibliographystyle{IEEEtran}
\bibliography{IEEEabrv,StringDefinitions,Papers,Books,Temp}









\end{document}

%% file: LatexInclusion/defmetric.tex
\newcommand{\bd}{\begin{description}}
\newcommand{\ed}{\end{description}}
\newcommand{\be}{\begin{enumerate}}
\newcommand{\ee}{\end{enumerate}}
\newcommand{\bi}{\begin{itemize}}
\newcommand{\ei}{\end{itemize}}
\newcommand{\bl}{\begin{list}}
\newcommand{\el}{\end{list}}
\newcommand{\bt}{\begin{tabbing}}
\newcommand{\et}{\end{tabbing}}

\definecolor{BLUE}{rgb}{0,0,1}

%% file: Figs/outage_r100.pgf
\begingroup%
\makeatletter%
\begin{pgfpicture}%
\pgfpathrectangle{\pgfpointorigin}{\pgfqpoint{6.000000in}{6.000000in}}%
\pgfusepath{use as bounding box, clip}%
\begin{pgfscope}%
\pgfsetbuttcap%
\pgfsetmiterjoin%
\definecolor{currentfill}{rgb}{1.000000,1.000000,1.000000}%
\pgfsetfillcolor{currentfill}%
\pgfsetlinewidth{0.000000pt}%
\definecolor{currentstroke}{rgb}{1.000000,1.000000,1.000000}%
\pgfsetstrokecolor{currentstroke}%
\pgfsetdash{}{0pt}%
\pgfpathmoveto{\pgfqpoint{0.000000in}{0.000000in}}%
\pgfpathlineto{\pgfqpoint{6.000000in}{0.000000in}}%
\pgfpathlineto{\pgfqpoint{6.000000in}{6.000000in}}%
\pgfpathlineto{\pgfqpoint{0.000000in}{6.000000in}}%
\pgfpathclose%
\pgfusepath{fill}%
\end{pgfscope}%
\begin{pgfscope}%
\pgfsetbuttcap%
\pgfsetmiterjoin%
\definecolor{currentfill}{rgb}{1.000000,1.000000,1.000000}%
\pgfsetfillcolor{currentfill}%
\pgfsetlinewidth{0.000000pt}%
\definecolor{currentstroke}{rgb}{0.000000,0.000000,0.000000}%
\pgfsetstrokecolor{currentstroke}%
\pgfsetstrokeopacity{0.000000}%
\pgfsetdash{}{0pt}%
\pgfpathmoveto{\pgfqpoint{0.822604in}{0.779809in}}%
\pgfpathlineto{\pgfqpoint{5.916667in}{0.779809in}}%
\pgfpathlineto{\pgfqpoint{5.916667in}{5.873872in}}%
\pgfpathlineto{\pgfqpoint{0.822604in}{5.873872in}}%
\pgfpathclose%
\pgfusepath{fill}%
\end{pgfscope}%
\begin{pgfscope}%
\pgfpathrectangle{\pgfqpoint{0.822604in}{0.779809in}}{\pgfqpoint{5.094063in}{5.094063in}}%
\pgfusepath{clip}%
\pgfsetrectcap%
\pgfsetroundjoin%
\pgfsetlinewidth{0.803000pt}%
\definecolor{currentstroke}{rgb}{0.690196,0.690196,0.690196}%
\pgfsetstrokecolor{currentstroke}%
\pgfsetdash{}{0pt}%
\pgfpathmoveto{\pgfqpoint{1.054152in}{0.779809in}}%
\pgfpathlineto{\pgfqpoint{1.054152in}{5.873872in}}%
\pgfusepath{stroke}%
\end{pgfscope}%
\begin{pgfscope}%
\pgfsetbuttcap%
\pgfsetroundjoin%
\definecolor{currentfill}{rgb}{0.000000,0.000000,0.000000}%
\pgfsetfillcolor{currentfill}%
\pgfsetlinewidth{0.803000pt}%
\definecolor{currentstroke}{rgb}{0.000000,0.000000,0.000000}%
\pgfsetstrokecolor{currentstroke}%
\pgfsetdash{}{0pt}%
\pgfsys@defobject{currentmarker}{\pgfqpoint{0.000000in}{-0.048611in}}{\pgfqpoint{0.000000in}{0.000000in}}{%
\pgfpathmoveto{\pgfqpoint{0.000000in}{0.000000in}}%
\pgfpathlineto{\pgfqpoint{0.000000in}{-0.048611in}}%
\pgfusepath{stroke,fill}%
}%
\begin{pgfscope}%
\pgfsys@transformshift{1.054152in}{0.779809in}%
\pgfsys@useobject{currentmarker}{}%
\end{pgfscope}%
\end{pgfscope}%
\begin{pgfscope}%
\definecolor{textcolor}{rgb}{0.000000,0.000000,0.000000}%
\pgfsetstrokecolor{textcolor}%
\pgfsetfillcolor{textcolor}%
\pgftext[x=1.054152in,y=0.682587in,,top]{\color{textcolor}\rmfamily\fontsize{18.000000}{21.600000}\selectfont \(\displaystyle {0}\)}%
\end{pgfscope}%
\begin{pgfscope}%
\pgfpathrectangle{\pgfqpoint{0.822604in}{0.779809in}}{\pgfqpoint{5.094063in}{5.094063in}}%
\pgfusepath{clip}%
\pgfsetrectcap%
\pgfsetroundjoin%
\pgfsetlinewidth{0.803000pt}%
\definecolor{currentstroke}{rgb}{0.690196,0.690196,0.690196}%
\pgfsetstrokecolor{currentstroke}%
\pgfsetdash{}{0pt}%
\pgfpathmoveto{\pgfqpoint{2.211894in}{0.779809in}}%
\pgfpathlineto{\pgfqpoint{2.211894in}{5.873872in}}%
\pgfusepath{stroke}%
\end{pgfscope}%
\begin{pgfscope}%
\pgfsetbuttcap%
\pgfsetroundjoin%
\definecolor{currentfill}{rgb}{0.000000,0.000000,0.000000}%
\pgfsetfillcolor{currentfill}%
\pgfsetlinewidth{0.803000pt}%
\definecolor{currentstroke}{rgb}{0.000000,0.000000,0.000000}%
\pgfsetstrokecolor{currentstroke}%
\pgfsetdash{}{0pt}%
\pgfsys@defobject{currentmarker}{\pgfqpoint{0.000000in}{-0.048611in}}{\pgfqpoint{0.000000in}{0.000000in}}{%
\pgfpathmoveto{\pgfqpoint{0.000000in}{0.000000in}}%
\pgfpathlineto{\pgfqpoint{0.000000in}{-0.048611in}}%
\pgfusepath{stroke,fill}%
}%
\begin{pgfscope}%
\pgfsys@transformshift{2.211894in}{0.779809in}%
\pgfsys@useobject{currentmarker}{}%
\end{pgfscope}%
\end{pgfscope}%
\begin{pgfscope}%
\definecolor{textcolor}{rgb}{0.000000,0.000000,0.000000}%
\pgfsetstrokecolor{textcolor}%
\pgfsetfillcolor{textcolor}%
\pgftext[x=2.211894in,y=0.682587in,,top]{\color{textcolor}\rmfamily\fontsize{18.000000}{21.600000}\selectfont \(\displaystyle {1}\)}%
\end{pgfscope}%
\begin{pgfscope}%
\pgfpathrectangle{\pgfqpoint{0.822604in}{0.779809in}}{\pgfqpoint{5.094063in}{5.094063in}}%
\pgfusepath{clip}%
\pgfsetrectcap%
\pgfsetroundjoin%
\pgfsetlinewidth{0.803000pt}%
\definecolor{currentstroke}{rgb}{0.690196,0.690196,0.690196}%
\pgfsetstrokecolor{currentstroke}%
\pgfsetdash{}{0pt}%
\pgfpathmoveto{\pgfqpoint{3.369635in}{0.779809in}}%
\pgfpathlineto{\pgfqpoint{3.369635in}{5.873872in}}%
\pgfusepath{stroke}%
\end{pgfscope}%
\begin{pgfscope}%
\pgfsetbuttcap%
\pgfsetroundjoin%
\definecolor{currentfill}{rgb}{0.000000,0.000000,0.000000}%
\pgfsetfillcolor{currentfill}%
\pgfsetlinewidth{0.803000pt}%
\definecolor{currentstroke}{rgb}{0.000000,0.000000,0.000000}%
\pgfsetstrokecolor{currentstroke}%
\pgfsetdash{}{0pt}%
\pgfsys@defobject{currentmarker}{\pgfqpoint{0.000000in}{-0.048611in}}{\pgfqpoint{0.000000in}{0.000000in}}{%
\pgfpathmoveto{\pgfqpoint{0.000000in}{0.000000in}}%
\pgfpathlineto{\pgfqpoint{0.000000in}{-0.048611in}}%
\pgfusepath{stroke,fill}%
}%
\begin{pgfscope}%
\pgfsys@transformshift{3.369635in}{0.779809in}%
\pgfsys@useobject{currentmarker}{}%
\end{pgfscope}%
\end{pgfscope}%
\begin{pgfscope}%
\definecolor{textcolor}{rgb}{0.000000,0.000000,0.000000}%
\pgfsetstrokecolor{textcolor}%
\pgfsetfillcolor{textcolor}%
\pgftext[x=3.369635in,y=0.682587in,,top]{\color{textcolor}\rmfamily\fontsize{18.000000}{21.600000}\selectfont \(\displaystyle {2}\)}%
\end{pgfscope}%
\begin{pgfscope}%
\pgfpathrectangle{\pgfqpoint{0.822604in}{0.779809in}}{\pgfqpoint{5.094063in}{5.094063in}}%
\pgfusepath{clip}%
\pgfsetrectcap%
\pgfsetroundjoin%
\pgfsetlinewidth{0.803000pt}%
\definecolor{currentstroke}{rgb}{0.690196,0.690196,0.690196}%
\pgfsetstrokecolor{currentstroke}%
\pgfsetdash{}{0pt}%
\pgfpathmoveto{\pgfqpoint{4.527377in}{0.779809in}}%
\pgfpathlineto{\pgfqpoint{4.527377in}{5.873872in}}%
\pgfusepath{stroke}%
\end{pgfscope}%
\begin{pgfscope}%
\pgfsetbuttcap%
\pgfsetroundjoin%
\definecolor{currentfill}{rgb}{0.000000,0.000000,0.000000}%
\pgfsetfillcolor{currentfill}%
\pgfsetlinewidth{0.803000pt}%
\definecolor{currentstroke}{rgb}{0.000000,0.000000,0.000000}%
\pgfsetstrokecolor{currentstroke}%
\pgfsetdash{}{0pt}%
\pgfsys@defobject{currentmarker}{\pgfqpoint{0.000000in}{-0.048611in}}{\pgfqpoint{0.000000in}{0.000000in}}{%
\pgfpathmoveto{\pgfqpoint{0.000000in}{0.000000in}}%
\pgfpathlineto{\pgfqpoint{0.000000in}{-0.048611in}}%
\pgfusepath{stroke,fill}%
}%
\begin{pgfscope}%
\pgfsys@transformshift{4.527377in}{0.779809in}%
\pgfsys@useobject{currentmarker}{}%
\end{pgfscope}%
\end{pgfscope}%
\begin{pgfscope}%
\definecolor{textcolor}{rgb}{0.000000,0.000000,0.000000}%
\pgfsetstrokecolor{textcolor}%
\pgfsetfillcolor{textcolor}%
\pgftext[x=4.527377in,y=0.682587in,,top]{\color{textcolor}\rmfamily\fontsize{18.000000}{21.600000}\selectfont \(\displaystyle {3}\)}%
\end{pgfscope}%
\begin{pgfscope}%
\pgfpathrectangle{\pgfqpoint{0.822604in}{0.779809in}}{\pgfqpoint{5.094063in}{5.094063in}}%
\pgfusepath{clip}%
\pgfsetrectcap%
\pgfsetroundjoin%
\pgfsetlinewidth{0.803000pt}%
\definecolor{currentstroke}{rgb}{0.690196,0.690196,0.690196}%
\pgfsetstrokecolor{currentstroke}%
\pgfsetdash{}{0pt}%
\pgfpathmoveto{\pgfqpoint{5.685118in}{0.779809in}}%
\pgfpathlineto{\pgfqpoint{5.685118in}{5.873872in}}%
\pgfusepath{stroke}%
\end{pgfscope}%
\begin{pgfscope}%
\pgfsetbuttcap%
\pgfsetroundjoin%
\definecolor{currentfill}{rgb}{0.000000,0.000000,0.000000}%
\pgfsetfillcolor{currentfill}%
\pgfsetlinewidth{0.803000pt}%
\definecolor{currentstroke}{rgb}{0.000000,0.000000,0.000000}%
\pgfsetstrokecolor{currentstroke}%
\pgfsetdash{}{0pt}%
\pgfsys@defobject{currentmarker}{\pgfqpoint{0.000000in}{-0.048611in}}{\pgfqpoint{0.000000in}{0.000000in}}{%
\pgfpathmoveto{\pgfqpoint{0.000000in}{0.000000in}}%
\pgfpathlineto{\pgfqpoint{0.000000in}{-0.048611in}}%
\pgfusepath{stroke,fill}%
}%
\begin{pgfscope}%
\pgfsys@transformshift{5.685118in}{0.779809in}%
\pgfsys@useobject{currentmarker}{}%
\end{pgfscope}%
\end{pgfscope}%
\begin{pgfscope}%
\definecolor{textcolor}{rgb}{0.000000,0.000000,0.000000}%
\pgfsetstrokecolor{textcolor}%
\pgfsetfillcolor{textcolor}%
\pgftext[x=5.685118in,y=0.682587in,,top]{\color{textcolor}\rmfamily\fontsize{18.000000}{21.600000}\selectfont \(\displaystyle {4}\)}%
\end{pgfscope}%
\begin{pgfscope}%
\definecolor{textcolor}{rgb}{0.000000,0.000000,0.000000}%
\pgfsetstrokecolor{textcolor}%
\pgfsetfillcolor{textcolor}%
\pgftext[x=3.36in,y=0.32in,,top]{\color{textcolor}\rmfamily\fontsize{20.000000}{24.000000}\selectfont \(\displaystyle e_{\text{th}}\) [m]}%
\end{pgfscope}%
\begin{pgfscope}%
\pgfpathrectangle{\pgfqpoint{0.822604in}{0.779809in}}{\pgfqpoint{5.094063in}{5.094063in}}%
\pgfusepath{clip}%
\pgfsetrectcap%
\pgfsetroundjoin%
\pgfsetlinewidth{0.803000pt}%
\definecolor{currentstroke}{rgb}{0.690196,0.690196,0.690196}%
\pgfsetstrokecolor{currentstroke}%
\pgfsetdash{}{0pt}%
\pgfpathmoveto{\pgfqpoint{0.822604in}{1.008445in}}%
\pgfpathlineto{\pgfqpoint{5.916667in}{1.008445in}}%
\pgfusepath{stroke}%
\end{pgfscope}%
\begin{pgfscope}%
\pgfsetbuttcap%
\pgfsetroundjoin%
\definecolor{currentfill}{rgb}{0.000000,0.000000,0.000000}%
\pgfsetfillcolor{currentfill}%
\pgfsetlinewidth{0.803000pt}%
\definecolor{currentstroke}{rgb}{0.000000,0.000000,0.000000}%
\pgfsetstrokecolor{currentstroke}%
\pgfsetdash{}{0pt}%
\pgfsys@defobject{currentmarker}{\pgfqpoint{-0.048611in}{0.000000in}}{\pgfqpoint{-0.000000in}{0.000000in}}{%
\pgfpathmoveto{\pgfqpoint{-0.000000in}{0.000000in}}%
\pgfpathlineto{\pgfqpoint{-0.048611in}{0.000000in}}%
\pgfusepath{stroke,fill}%
}%
\begin{pgfscope}%
\pgfsys@transformshift{0.822604in}{1.008445in}%
\pgfsys@useobject{currentmarker}{}%
\end{pgfscope}%
\end{pgfscope}%
\begin{pgfscope}%
\definecolor{textcolor}{rgb}{0.000000,0.000000,0.000000}%
\pgfsetstrokecolor{textcolor}%
\pgfsetfillcolor{textcolor}%
\pgftext[x=0.439968in, y=0.925112in, left, base]{\color{textcolor}\rmfamily\fontsize{18.000000}{21.600000}\selectfont \(\displaystyle {0.0}\)}%
\end{pgfscope}%
\begin{pgfscope}%
\pgfpathrectangle{\pgfqpoint{0.822604in}{0.779809in}}{\pgfqpoint{5.094063in}{5.094063in}}%
\pgfusepath{clip}%
\pgfsetrectcap%
\pgfsetroundjoin%
\pgfsetlinewidth{0.803000pt}%
\definecolor{currentstroke}{rgb}{0.690196,0.690196,0.690196}%
\pgfsetstrokecolor{currentstroke}%
\pgfsetdash{}{0pt}%
\pgfpathmoveto{\pgfqpoint{0.822604in}{1.935221in}}%
\pgfpathlineto{\pgfqpoint{5.916667in}{1.935221in}}%
\pgfusepath{stroke}%
\end{pgfscope}%
\begin{pgfscope}%
\pgfsetbuttcap%
\pgfsetroundjoin%
\definecolor{currentfill}{rgb}{0.000000,0.000000,0.000000}%
\pgfsetfillcolor{currentfill}%
\pgfsetlinewidth{0.803000pt}%
\definecolor{currentstroke}{rgb}{0.000000,0.000000,0.000000}%
\pgfsetstrokecolor{currentstroke}%
\pgfsetdash{}{0pt}%
\pgfsys@defobject{currentmarker}{\pgfqpoint{-0.048611in}{0.000000in}}{\pgfqpoint{-0.000000in}{0.000000in}}{%
\pgfpathmoveto{\pgfqpoint{-0.000000in}{0.000000in}}%
\pgfpathlineto{\pgfqpoint{-0.048611in}{0.000000in}}%
\pgfusepath{stroke,fill}%
}%
\begin{pgfscope}%
\pgfsys@transformshift{0.822604in}{1.935221in}%
\pgfsys@useobject{currentmarker}{}%
\end{pgfscope}%
\end{pgfscope}%
\begin{pgfscope}%
\definecolor{textcolor}{rgb}{0.000000,0.000000,0.000000}%
\pgfsetstrokecolor{textcolor}%
\pgfsetfillcolor{textcolor}%
\pgftext[x=0.439968in, y=1.851888in, left, base]{\color{textcolor}\rmfamily\fontsize{18.000000}{21.600000}\selectfont \(\displaystyle {0.2}\)}%
\end{pgfscope}%
\begin{pgfscope}%
\pgfpathrectangle{\pgfqpoint{0.822604in}{0.779809in}}{\pgfqpoint{5.094063in}{5.094063in}}%
\pgfusepath{clip}%
\pgfsetrectcap%
\pgfsetroundjoin%
\pgfsetlinewidth{0.803000pt}%
\definecolor{currentstroke}{rgb}{0.690196,0.690196,0.690196}%
\pgfsetstrokecolor{currentstroke}%
\pgfsetdash{}{0pt}%
\pgfpathmoveto{\pgfqpoint{0.822604in}{2.861997in}}%
\pgfpathlineto{\pgfqpoint{5.916667in}{2.861997in}}%
\pgfusepath{stroke}%
\end{pgfscope}%
\begin{pgfscope}%
\pgfsetbuttcap%
\pgfsetroundjoin%
\definecolor{currentfill}{rgb}{0.000000,0.000000,0.000000}%
\pgfsetfillcolor{currentfill}%
\pgfsetlinewidth{0.803000pt}%
\definecolor{currentstroke}{rgb}{0.000000,0.000000,0.000000}%
\pgfsetstrokecolor{currentstroke}%
\pgfsetdash{}{0pt}%
\pgfsys@defobject{currentmarker}{\pgfqpoint{-0.048611in}{0.000000in}}{\pgfqpoint{-0.000000in}{0.000000in}}{%
\pgfpathmoveto{\pgfqpoint{-0.000000in}{0.000000in}}%
\pgfpathlineto{\pgfqpoint{-0.048611in}{0.000000in}}%
\pgfusepath{stroke,fill}%
}%
\begin{pgfscope}%
\pgfsys@transformshift{0.822604in}{2.861997in}%
\pgfsys@useobject{currentmarker}{}%
\end{pgfscope}%
\end{pgfscope}%
\begin{pgfscope}%
\definecolor{textcolor}{rgb}{0.000000,0.000000,0.000000}%
\pgfsetstrokecolor{textcolor}%
\pgfsetfillcolor{textcolor}%
\pgftext[x=0.439968in, y=2.778663in, left, base]{\color{textcolor}\rmfamily\fontsize{18.000000}{21.600000}\selectfont \(\displaystyle {0.4}\)}%
\end{pgfscope}%
\begin{pgfscope}%
\pgfpathrectangle{\pgfqpoint{0.822604in}{0.779809in}}{\pgfqpoint{5.094063in}{5.094063in}}%
\pgfusepath{clip}%
\pgfsetrectcap%
\pgfsetroundjoin%
\pgfsetlinewidth{0.803000pt}%
\definecolor{currentstroke}{rgb}{0.690196,0.690196,0.690196}%
\pgfsetstrokecolor{currentstroke}%
\pgfsetdash{}{0pt}%
\pgfpathmoveto{\pgfqpoint{0.822604in}{3.788773in}}%
\pgfpathlineto{\pgfqpoint{5.916667in}{3.788773in}}%
\pgfusepath{stroke}%
\end{pgfscope}%
\begin{pgfscope}%
\pgfsetbuttcap%
\pgfsetroundjoin%
\definecolor{currentfill}{rgb}{0.000000,0.000000,0.000000}%
\pgfsetfillcolor{currentfill}%
\pgfsetlinewidth{0.803000pt}%
\definecolor{currentstroke}{rgb}{0.000000,0.000000,0.000000}%
\pgfsetstrokecolor{currentstroke}%
\pgfsetdash{}{0pt}%
\pgfsys@defobject{currentmarker}{\pgfqpoint{-0.048611in}{0.000000in}}{\pgfqpoint{-0.000000in}{0.000000in}}{%
\pgfpathmoveto{\pgfqpoint{-0.000000in}{0.000000in}}%
\pgfpathlineto{\pgfqpoint{-0.048611in}{0.000000in}}%
\pgfusepath{stroke,fill}%
}%
\begin{pgfscope}%
\pgfsys@transformshift{0.822604in}{3.788773in}%
\pgfsys@useobject{currentmarker}{}%
\end{pgfscope}%
\end{pgfscope}%
\begin{pgfscope}%
\definecolor{textcolor}{rgb}{0.000000,0.000000,0.000000}%
\pgfsetstrokecolor{textcolor}%
\pgfsetfillcolor{textcolor}%
\pgftext[x=0.439968in, y=3.705439in, left, base]{\color{textcolor}\rmfamily\fontsize{18.000000}{21.600000}\selectfont \(\displaystyle {0.6}\)}%
\end{pgfscope}%
\begin{pgfscope}%
\pgfpathrectangle{\pgfqpoint{0.822604in}{0.779809in}}{\pgfqpoint{5.094063in}{5.094063in}}%
\pgfusepath{clip}%
\pgfsetrectcap%
\pgfsetroundjoin%
\pgfsetlinewidth{0.803000pt}%
\definecolor{currentstroke}{rgb}{0.690196,0.690196,0.690196}%
\pgfsetstrokecolor{currentstroke}%
\pgfsetdash{}{0pt}%
\pgfpathmoveto{\pgfqpoint{0.822604in}{4.715548in}}%
\pgfpathlineto{\pgfqpoint{5.916667in}{4.715548in}}%
\pgfusepath{stroke}%
\end{pgfscope}%
\begin{pgfscope}%
\pgfsetbuttcap%
\pgfsetroundjoin%
\definecolor{currentfill}{rgb}{0.000000,0.000000,0.000000}%
\pgfsetfillcolor{currentfill}%
\pgfsetlinewidth{0.803000pt}%
\definecolor{currentstroke}{rgb}{0.000000,0.000000,0.000000}%
\pgfsetstrokecolor{currentstroke}%
\pgfsetdash{}{0pt}%
\pgfsys@defobject{currentmarker}{\pgfqpoint{-0.048611in}{0.000000in}}{\pgfqpoint{-0.000000in}{0.000000in}}{%
\pgfpathmoveto{\pgfqpoint{-0.000000in}{0.000000in}}%
\pgfpathlineto{\pgfqpoint{-0.048611in}{0.000000in}}%
\pgfusepath{stroke,fill}%
}%
\begin{pgfscope}%
\pgfsys@transformshift{0.822604in}{4.715548in}%
\pgfsys@useobject{currentmarker}{}%
\end{pgfscope}%
\end{pgfscope}%
\begin{pgfscope}%
\definecolor{textcolor}{rgb}{0.000000,0.000000,0.000000}%
\pgfsetstrokecolor{textcolor}%
\pgfsetfillcolor{textcolor}%
\pgftext[x=0.439968in, y=4.632215in, left, base]{\color{textcolor}\rmfamily\fontsize{18.000000}{21.600000}\selectfont \(\displaystyle {0.8}\)}%
\end{pgfscope}%
\begin{pgfscope}%
\pgfpathrectangle{\pgfqpoint{0.822604in}{0.779809in}}{\pgfqpoint{5.094063in}{5.094063in}}%
\pgfusepath{clip}%
\pgfsetrectcap%
\pgfsetroundjoin%
\pgfsetlinewidth{0.803000pt}%
\definecolor{currentstroke}{rgb}{0.690196,0.690196,0.690196}%
\pgfsetstrokecolor{currentstroke}%
\pgfsetdash{}{0pt}%
\pgfpathmoveto{\pgfqpoint{0.822604in}{5.642324in}}%
\pgfpathlineto{\pgfqpoint{5.916667in}{5.642324in}}%
\pgfusepath{stroke}%
\end{pgfscope}%
\begin{pgfscope}%
\pgfsetbuttcap%
\pgfsetroundjoin%
\definecolor{currentfill}{rgb}{0.000000,0.000000,0.000000}%
\pgfsetfillcolor{currentfill}%
\pgfsetlinewidth{0.803000pt}%
\definecolor{currentstroke}{rgb}{0.000000,0.000000,0.000000}%
\pgfsetstrokecolor{currentstroke}%
\pgfsetdash{}{0pt}%
\pgfsys@defobject{currentmarker}{\pgfqpoint{-0.048611in}{0.000000in}}{\pgfqpoint{-0.000000in}{0.000000in}}{%
\pgfpathmoveto{\pgfqpoint{-0.000000in}{0.000000in}}%
\pgfpathlineto{\pgfqpoint{-0.048611in}{0.000000in}}%
\pgfusepath{stroke,fill}%
}%
\begin{pgfscope}%
\pgfsys@transformshift{0.822604in}{5.642324in}%
\pgfsys@useobject{currentmarker}{}%
\end{pgfscope}%
\end{pgfscope}%
\begin{pgfscope}%
\definecolor{textcolor}{rgb}{0.000000,0.000000,0.000000}%
\pgfsetstrokecolor{textcolor}%
\pgfsetfillcolor{textcolor}%
\pgftext[x=0.439968in, y=5.558991in, left, base]{\color{textcolor}\rmfamily\fontsize{18.000000}{21.600000}\selectfont \(\displaystyle {1.0}\)}%
\end{pgfscope}%
\begin{pgfscope}%
\definecolor{textcolor}{rgb}{0.000000,0.000000,0.000000}%
\pgfsetstrokecolor{textcolor}%
\pgfsetfillcolor{textcolor}%
\pgftext[x=0.32in,y=3.30in,,bottom,rotate=90.000000]{\color{textcolor}\rmfamily\fontsize{20.000000}{24.000000}\selectfont \(\displaystyle \hat{P}_{\text{out}}\)}%
\end{pgfscope}%
\begin{pgfscope}%
\pgfpathrectangle{\pgfqpoint{0.822604in}{0.779809in}}{\pgfqpoint{5.094063in}{5.094063in}}%
\pgfusepath{clip}%
\pgfsetrectcap%
\pgfsetroundjoin%
\pgfsetlinewidth{1.505625pt}%
\definecolor{currentstroke}{rgb}{0.000000,0.000000,1.000000}%
\pgfsetstrokecolor{currentstroke}%
\pgfsetdash{}{0pt}%
\pgfpathmoveto{\pgfqpoint{1.054152in}{5.642324in}}%
\pgfpathlineto{\pgfqpoint{1.065729in}{5.641249in}}%
\pgfpathlineto{\pgfqpoint{1.077307in}{5.638005in}}%
\pgfpathlineto{\pgfqpoint{1.088884in}{5.632530in}}%
\pgfpathlineto{\pgfqpoint{1.100462in}{5.624972in}}%
\pgfpathlineto{\pgfqpoint{1.112039in}{5.615276in}}%
\pgfpathlineto{\pgfqpoint{1.123617in}{5.603425in}}%
\pgfpathlineto{\pgfqpoint{1.146771in}{5.573015in}}%
\pgfpathlineto{\pgfqpoint{1.169926in}{5.533616in}}%
\pgfpathlineto{\pgfqpoint{1.193081in}{5.486990in}}%
\pgfpathlineto{\pgfqpoint{1.216236in}{5.432609in}}%
\pgfpathlineto{\pgfqpoint{1.239391in}{5.370030in}}%
\pgfpathlineto{\pgfqpoint{1.262546in}{5.300992in}}%
\pgfpathlineto{\pgfqpoint{1.285700in}{5.225073in}}%
\pgfpathlineto{\pgfqpoint{1.308855in}{5.143005in}}%
\pgfpathlineto{\pgfqpoint{1.343587in}{5.010300in}}%
\pgfpathlineto{\pgfqpoint{1.378320in}{4.866509in}}%
\pgfpathlineto{\pgfqpoint{1.413052in}{4.713683in}}%
\pgfpathlineto{\pgfqpoint{1.459362in}{4.497501in}}%
\pgfpathlineto{\pgfqpoint{1.528826in}{4.161040in}}%
\pgfpathlineto{\pgfqpoint{1.644600in}{3.596828in}}%
\pgfpathlineto{\pgfqpoint{1.702487in}{3.326879in}}%
\pgfpathlineto{\pgfqpoint{1.748797in}{3.122050in}}%
\pgfpathlineto{\pgfqpoint{1.783529in}{2.974978in}}%
\pgfpathlineto{\pgfqpoint{1.829839in}{2.790547in}}%
\pgfpathlineto{\pgfqpoint{1.864571in}{2.660877in}}%
\pgfpathlineto{\pgfqpoint{1.899303in}{2.538026in}}%
\pgfpathlineto{\pgfqpoint{1.945613in}{2.387138in}}%
\pgfpathlineto{\pgfqpoint{1.980345in}{2.281634in}}%
\pgfpathlineto{\pgfqpoint{2.015078in}{2.183810in}}%
\pgfpathlineto{\pgfqpoint{2.049810in}{2.092504in}}%
\pgfpathlineto{\pgfqpoint{2.084542in}{2.008487in}}%
\pgfpathlineto{\pgfqpoint{2.130852in}{1.906452in}}%
\pgfpathlineto{\pgfqpoint{2.154007in}{1.859844in}}%
\pgfpathlineto{\pgfqpoint{2.188739in}{1.794265in}}%
\pgfpathlineto{\pgfqpoint{2.223471in}{1.734243in}}%
\pgfpathlineto{\pgfqpoint{2.258203in}{1.678898in}}%
\pgfpathlineto{\pgfqpoint{2.292936in}{1.629026in}}%
\pgfpathlineto{\pgfqpoint{2.327668in}{1.583299in}}%
\pgfpathlineto{\pgfqpoint{2.362400in}{1.540934in}}%
\pgfpathlineto{\pgfqpoint{2.397132in}{1.502178in}}%
\pgfpathlineto{\pgfqpoint{2.431865in}{1.466886in}}%
\pgfpathlineto{\pgfqpoint{2.478174in}{1.424382in}}%
\pgfpathlineto{\pgfqpoint{2.512906in}{1.395383in}}%
\pgfpathlineto{\pgfqpoint{2.547639in}{1.369070in}}%
\pgfpathlineto{\pgfqpoint{2.582371in}{1.344893in}}%
\pgfpathlineto{\pgfqpoint{2.628681in}{1.315736in}}%
\pgfpathlineto{\pgfqpoint{2.686568in}{1.284277in}}%
\pgfpathlineto{\pgfqpoint{2.721300in}{1.267187in}}%
\pgfpathlineto{\pgfqpoint{2.767610in}{1.246643in}}%
\pgfpathlineto{\pgfqpoint{2.837074in}{1.219743in}}%
\pgfpathlineto{\pgfqpoint{2.918116in}{1.193200in}}%
\pgfpathlineto{\pgfqpoint{2.976003in}{1.176745in}}%
\pgfpathlineto{\pgfqpoint{3.033890in}{1.161792in}}%
\pgfpathlineto{\pgfqpoint{3.103355in}{1.146211in}}%
\pgfpathlineto{\pgfqpoint{3.184397in}{1.130446in}}%
\pgfpathlineto{\pgfqpoint{3.288593in}{1.112631in}}%
\pgfpathlineto{\pgfqpoint{3.369635in}{1.100979in}}%
\pgfpathlineto{\pgfqpoint{3.496987in}{1.085479in}}%
\pgfpathlineto{\pgfqpoint{3.693803in}{1.066807in}}%
\pgfpathlineto{\pgfqpoint{3.936929in}{1.050488in}}%
\pgfpathlineto{\pgfqpoint{4.087435in}{1.043049in}}%
\pgfpathlineto{\pgfqpoint{4.353716in}{1.032949in}}%
\pgfpathlineto{\pgfqpoint{4.747348in}{1.023130in}}%
\pgfpathlineto{\pgfqpoint{5.071515in}{1.018450in}}%
\pgfpathlineto{\pgfqpoint{5.523035in}{1.014395in}}%
\pgfpathlineto{\pgfqpoint{5.685118in}{1.013401in}}%
\pgfpathlineto{\pgfqpoint{5.685118in}{1.013401in}}%
\pgfusepath{stroke}%
\end{pgfscope}%
\begin{pgfscope}%
\pgfpathrectangle{\pgfqpoint{0.822604in}{0.779809in}}{\pgfqpoint{5.094063in}{5.094063in}}%
\pgfusepath{clip}%
\pgfsetrectcap%
\pgfsetroundjoin%
\pgfsetlinewidth{1.505625pt}%
\definecolor{currentstroke}{rgb}{1.000000,0.300000,0.300000}%
\pgfsetstrokecolor{currentstroke}%
\pgfsetdash{}{0pt}%
\pgfpathmoveto{\pgfqpoint{1.054152in}{5.642324in}}%
\pgfpathlineto{\pgfqpoint{1.065729in}{5.640336in}}%
\pgfpathlineto{\pgfqpoint{1.077307in}{5.634428in}}%
\pgfpathlineto{\pgfqpoint{1.088884in}{5.624391in}}%
\pgfpathlineto{\pgfqpoint{1.100462in}{5.610438in}}%
\pgfpathlineto{\pgfqpoint{1.112039in}{5.592992in}}%
\pgfpathlineto{\pgfqpoint{1.123617in}{5.572060in}}%
\pgfpathlineto{\pgfqpoint{1.135194in}{5.546884in}}%
\pgfpathlineto{\pgfqpoint{1.146771in}{5.518115in}}%
\pgfpathlineto{\pgfqpoint{1.158349in}{5.485838in}}%
\pgfpathlineto{\pgfqpoint{1.181504in}{5.411691in}}%
\pgfpathlineto{\pgfqpoint{1.204658in}{5.323597in}}%
\pgfpathlineto{\pgfqpoint{1.227813in}{5.222747in}}%
\pgfpathlineto{\pgfqpoint{1.250968in}{5.111889in}}%
\pgfpathlineto{\pgfqpoint{1.274123in}{4.990861in}}%
\pgfpathlineto{\pgfqpoint{1.308855in}{4.793620in}}%
\pgfpathlineto{\pgfqpoint{1.343587in}{4.581594in}}%
\pgfpathlineto{\pgfqpoint{1.389897in}{4.281194in}}%
\pgfpathlineto{\pgfqpoint{1.540404in}{3.292303in}}%
\pgfpathlineto{\pgfqpoint{1.586713in}{3.011532in}}%
\pgfpathlineto{\pgfqpoint{1.621445in}{2.815711in}}%
\pgfpathlineto{\pgfqpoint{1.667755in}{2.571749in}}%
\pgfpathlineto{\pgfqpoint{1.702487in}{2.404631in}}%
\pgfpathlineto{\pgfqpoint{1.725642in}{2.300917in}}%
\pgfpathlineto{\pgfqpoint{1.760374in}{2.158194in}}%
\pgfpathlineto{\pgfqpoint{1.795107in}{2.029076in}}%
\pgfpathlineto{\pgfqpoint{1.829839in}{1.912770in}}%
\pgfpathlineto{\pgfqpoint{1.852994in}{1.841874in}}%
\pgfpathlineto{\pgfqpoint{1.887726in}{1.746048in}}%
\pgfpathlineto{\pgfqpoint{1.910881in}{1.688136in}}%
\pgfpathlineto{\pgfqpoint{1.934036in}{1.634461in}}%
\pgfpathlineto{\pgfqpoint{1.957190in}{1.585354in}}%
\pgfpathlineto{\pgfqpoint{1.980345in}{1.540815in}}%
\pgfpathlineto{\pgfqpoint{2.003500in}{1.499490in}}%
\pgfpathlineto{\pgfqpoint{2.038232in}{1.444528in}}%
\pgfpathlineto{\pgfqpoint{2.061387in}{1.411162in}}%
\pgfpathlineto{\pgfqpoint{2.084542in}{1.380898in}}%
\pgfpathlineto{\pgfqpoint{2.107697in}{1.353403in}}%
\pgfpathlineto{\pgfqpoint{2.130852in}{1.328336in}}%
\pgfpathlineto{\pgfqpoint{2.165584in}{1.294775in}}%
\pgfpathlineto{\pgfqpoint{2.200316in}{1.266101in}}%
\pgfpathlineto{\pgfqpoint{2.223471in}{1.248619in}}%
\pgfpathlineto{\pgfqpoint{2.269781in}{1.217792in}}%
\pgfpathlineto{\pgfqpoint{2.304513in}{1.198590in}}%
\pgfpathlineto{\pgfqpoint{2.350823in}{1.176266in}}%
\pgfpathlineto{\pgfqpoint{2.385555in}{1.161822in}}%
\pgfpathlineto{\pgfqpoint{2.431865in}{1.145369in}}%
\pgfpathlineto{\pgfqpoint{2.478174in}{1.131208in}}%
\pgfpathlineto{\pgfqpoint{2.524484in}{1.119251in}}%
\pgfpathlineto{\pgfqpoint{2.570794in}{1.108984in}}%
\pgfpathlineto{\pgfqpoint{2.663413in}{1.092717in}}%
\pgfpathlineto{\pgfqpoint{2.732877in}{1.082817in}}%
\pgfpathlineto{\pgfqpoint{2.813919in}{1.073197in}}%
\pgfpathlineto{\pgfqpoint{2.894961in}{1.065437in}}%
\pgfpathlineto{\pgfqpoint{2.987580in}{1.058016in}}%
\pgfpathlineto{\pgfqpoint{3.080200in}{1.051740in}}%
\pgfpathlineto{\pgfqpoint{3.242284in}{1.043169in}}%
\pgfpathlineto{\pgfqpoint{3.578029in}{1.031492in}}%
\pgfpathlineto{\pgfqpoint{3.925351in}{1.023971in}}%
\pgfpathlineto{\pgfqpoint{4.261096in}{1.019231in}}%
\pgfpathlineto{\pgfqpoint{4.851544in}{1.014409in}}%
\pgfpathlineto{\pgfqpoint{5.256754in}{1.012562in}}%
\pgfpathlineto{\pgfqpoint{5.685118in}{1.011358in}}%
\pgfpathlineto{\pgfqpoint{5.685118in}{1.011358in}}%
\pgfusepath{stroke}%
\end{pgfscope}%
\begin{pgfscope}%
\pgfsetrectcap%
\pgfsetmiterjoin%
\pgfsetlinewidth{0.803000pt}%
\definecolor{currentstroke}{rgb}{0.000000,0.000000,0.000000}%
\pgfsetstrokecolor{currentstroke}%
\pgfsetdash{}{0pt}%
\pgfpathmoveto{\pgfqpoint{0.822604in}{0.779809in}}%
\pgfpathlineto{\pgfqpoint{0.822604in}{5.873872in}}%
\pgfusepath{stroke}%
\end{pgfscope}%
\begin{pgfscope}%
\pgfsetrectcap%
\pgfsetmiterjoin%
\pgfsetlinewidth{0.803000pt}%
\definecolor{currentstroke}{rgb}{0.000000,0.000000,0.000000}%
\pgfsetstrokecolor{currentstroke}%
\pgfsetdash{}{0pt}%
\pgfpathmoveto{\pgfqpoint{5.916667in}{0.779809in}}%
\pgfpathlineto{\pgfqpoint{5.916667in}{5.873872in}}%
\pgfusepath{stroke}%
\end{pgfscope}%
\begin{pgfscope}%
\pgfsetrectcap%
\pgfsetmiterjoin%
\pgfsetlinewidth{0.803000pt}%
\definecolor{currentstroke}{rgb}{0.000000,0.000000,0.000000}%
\pgfsetstrokecolor{currentstroke}%
\pgfsetdash{}{0pt}%
\pgfpathmoveto{\pgfqpoint{0.822604in}{0.779809in}}%
\pgfpathlineto{\pgfqpoint{5.916667in}{0.779809in}}%
\pgfusepath{stroke}%
\end{pgfscope}%
\begin{pgfscope}%
\pgfsetrectcap%
\pgfsetmiterjoin%
\pgfsetlinewidth{0.803000pt}%
\definecolor{currentstroke}{rgb}{0.000000,0.000000,0.000000}%
\pgfsetstrokecolor{currentstroke}%
\pgfsetdash{}{0pt}%
\pgfpathmoveto{\pgfqpoint{0.822604in}{5.873872in}}%
\pgfpathlineto{\pgfqpoint{5.916667in}{5.873872in}}%
\pgfusepath{stroke}%
\end{pgfscope}%
\begin{pgfscope}%
\pgfsetbuttcap%
\pgfsetmiterjoin%
\definecolor{currentfill}{rgb}{1.000000,1.000000,1.000000}%
\pgfsetfillcolor{currentfill}%
\pgfsetfillopacity{0.800000}%
\pgfsetlinewidth{1.003750pt}%
\definecolor{currentstroke}{rgb}{0.800000,0.800000,0.800000}%
\pgfsetstrokecolor{currentstroke}%
\pgfsetstrokeopacity{0.800000}%
\pgfsetdash{}{0pt}%
\pgfpathmoveto{\pgfqpoint{2.722645in}{4.830249in}}%
\pgfpathlineto{\pgfqpoint{5.722222in}{4.830249in}}%
\pgfpathquadraticcurveto{\pgfqpoint{5.777778in}{4.830249in}}{\pgfqpoint{5.777778in}{4.885805in}}%
\pgfpathlineto{\pgfqpoint{5.777778in}{5.679428in}}%
\pgfpathquadraticcurveto{\pgfqpoint{5.777778in}{5.734983in}}{\pgfqpoint{5.722222in}{5.734983in}}%
\pgfpathlineto{\pgfqpoint{2.722645in}{5.734983in}}%
\pgfpathquadraticcurveto{\pgfqpoint{2.667090in}{5.734983in}}{\pgfqpoint{2.667090in}{5.679428in}}%
\pgfpathlineto{\pgfqpoint{2.667090in}{4.885805in}}%
\pgfpathquadraticcurveto{\pgfqpoint{2.667090in}{4.830249in}}{\pgfqpoint{2.722645in}{4.830249in}}%
\pgfpathclose%
\pgfusepath{stroke,fill}%
\end{pgfscope}%
\begin{pgfscope}%
\pgfsetrectcap%
\pgfsetroundjoin%
\pgfsetlinewidth{1.505625pt}%
\definecolor{currentstroke}{rgb}{0.000000,0.000000,1.000000}%
\pgfsetstrokecolor{currentstroke}%
\pgfsetdash{}{0pt}%
\pgfpathmoveto{\pgfqpoint{2.778201in}{5.521056in}}%
\pgfpathlineto{\pgfqpoint{3.333756in}{5.521056in}}%
\pgfusepath{stroke}%
\end{pgfscope}%
\begin{pgfscope}%
\definecolor{textcolor}{rgb}{0.000000,0.000000,0.000000}%
\pgfsetstrokecolor{textcolor}%
\pgfsetfillcolor{textcolor}%
\pgftext[x=3.555979in,y=5.423834in,left,base]{\color{textcolor}\rmfamily\fontsize{20.000000}{24.000000}\selectfont BP}%
\end{pgfscope}%
\begin{pgfscope}%
\pgfsetrectcap%
\pgfsetroundjoin%
\pgfsetlinewidth{1.505625pt}%
\definecolor{currentstroke}{rgb}{1.000000,0.300000,0.300000}%
\pgfsetstrokecolor{currentstroke}%
\pgfsetdash{}{0pt}%
\pgfpathmoveto{\pgfqpoint{2.778201in}{5.110356in}}%
\pgfpathlineto{\pgfqpoint{3.333756in}{5.110356in}}%
\pgfusepath{stroke}%
\end{pgfscope}%
\begin{pgfscope}%
\definecolor{textcolor}{rgb}{0.000000,0.000000,0.000000}%
\pgfsetstrokecolor{textcolor}%
\pgfsetfillcolor{textcolor}%
\pgftext[x=3.555979in,y=5.013133in,left,base]{\color{textcolor}\rmfamily\fontsize{20.000000}{24.000000}\selectfont NEBP (proposed)}%
\end{pgfscope}%
\end{pgfpicture}%
\makeatother%
\endgroup%

%% file: Figs/confidence_percent_twosided_r100.pgf
\begingroup%
\makeatletter%
\begin{pgfpicture}%
\pgfpathrectangle{\pgfpointorigin}{\pgfqpoint{6.000000in}{6.000000in}}%
\pgfusepath{use as bounding box, clip}%
\begin{pgfscope}%
\pgfsetbuttcap%
\pgfsetmiterjoin%
\definecolor{currentfill}{rgb}{1.000000,1.000000,1.000000}%
\pgfsetfillcolor{currentfill}%
\pgfsetlinewidth{0.000000pt}%
\definecolor{currentstroke}{rgb}{1.000000,1.000000,1.000000}%
\pgfsetstrokecolor{currentstroke}%
\pgfsetdash{}{0pt}%
\pgfpathmoveto{\pgfqpoint{0.000000in}{0.000000in}}%
\pgfpathlineto{\pgfqpoint{6.000000in}{0.000000in}}%
\pgfpathlineto{\pgfqpoint{6.000000in}{6.000000in}}%
\pgfpathlineto{\pgfqpoint{0.000000in}{6.000000in}}%
\pgfpathclose%
\pgfusepath{fill}%
\end{pgfscope}%
\begin{pgfscope}%
\pgfsetbuttcap%
\pgfsetmiterjoin%
\definecolor{currentfill}{rgb}{1.000000,1.000000,1.000000}%
\pgfsetfillcolor{currentfill}%
\pgfsetlinewidth{0.000000pt}%
\definecolor{currentstroke}{rgb}{0.000000,0.000000,0.000000}%
\pgfsetstrokecolor{currentstroke}%
\pgfsetstrokeopacity{0.000000}%
\pgfsetdash{}{0pt}%
\pgfpathmoveto{\pgfqpoint{0.794080in}{0.749804in}}%
\pgfpathlineto{\pgfqpoint{5.916667in}{0.749804in}}%
\pgfpathlineto{\pgfqpoint{5.916667in}{5.872390in}}%
\pgfpathlineto{\pgfqpoint{0.794080in}{5.872390in}}%
\pgfpathclose%
\pgfusepath{fill}%
\end{pgfscope}%
\begin{pgfscope}%
\pgfpathrectangle{\pgfqpoint{0.794080in}{0.749804in}}{\pgfqpoint{5.122587in}{5.122587in}}%
\pgfusepath{clip}%
\pgfsetrectcap%
\pgfsetroundjoin%
\pgfsetlinewidth{0.803000pt}%
\definecolor{currentstroke}{rgb}{0.690196,0.690196,0.690196}%
\pgfsetstrokecolor{currentstroke}%
\pgfsetdash{}{0pt}%
\pgfpathmoveto{\pgfqpoint{0.983805in}{0.749804in}}%
\pgfpathlineto{\pgfqpoint{0.983805in}{5.872390in}}%
\pgfusepath{stroke}%
\end{pgfscope}%
\begin{pgfscope}%
\pgfsetbuttcap%
\pgfsetroundjoin%
\definecolor{currentfill}{rgb}{0.000000,0.000000,0.000000}%
\pgfsetfillcolor{currentfill}%
\pgfsetlinewidth{0.803000pt}%
\definecolor{currentstroke}{rgb}{0.000000,0.000000,0.000000}%
\pgfsetstrokecolor{currentstroke}%
\pgfsetdash{}{0pt}%
\pgfsys@defobject{currentmarker}{\pgfqpoint{0.000000in}{-0.048611in}}{\pgfqpoint{0.000000in}{0.000000in}}{%
\pgfpathmoveto{\pgfqpoint{0.000000in}{0.000000in}}%
\pgfpathlineto{\pgfqpoint{0.000000in}{-0.048611in}}%
\pgfusepath{stroke,fill}%
}%
\begin{pgfscope}%
\pgfsys@transformshift{0.983805in}{0.749804in}%
\pgfsys@useobject{currentmarker}{}%
\end{pgfscope}%
\end{pgfscope}%
\begin{pgfscope}%
\definecolor{textcolor}{rgb}{0.000000,0.000000,0.000000}%
\pgfsetstrokecolor{textcolor}%
\pgfsetfillcolor{textcolor}%
\pgftext[x=0.983805in,y=0.652581in,,top]{\color{textcolor}\rmfamily\fontsize{18.000000}{21.600000}\selectfont \(\displaystyle {0.0}\)}%
\end{pgfscope}%
\begin{pgfscope}%
\pgfpathrectangle{\pgfqpoint{0.794080in}{0.749804in}}{\pgfqpoint{5.122587in}{5.122587in}}%
\pgfusepath{clip}%
\pgfsetrectcap%
\pgfsetroundjoin%
\pgfsetlinewidth{0.803000pt}%
\definecolor{currentstroke}{rgb}{0.690196,0.690196,0.690196}%
\pgfsetstrokecolor{currentstroke}%
\pgfsetdash{}{0pt}%
\pgfpathmoveto{\pgfqpoint{1.932432in}{0.749804in}}%
\pgfpathlineto{\pgfqpoint{1.932432in}{5.872390in}}%
\pgfusepath{stroke}%
\end{pgfscope}%
\begin{pgfscope}%
\pgfsetbuttcap%
\pgfsetroundjoin%
\definecolor{currentfill}{rgb}{0.000000,0.000000,0.000000}%
\pgfsetfillcolor{currentfill}%
\pgfsetlinewidth{0.803000pt}%
\definecolor{currentstroke}{rgb}{0.000000,0.000000,0.000000}%
\pgfsetstrokecolor{currentstroke}%
\pgfsetdash{}{0pt}%
\pgfsys@defobject{currentmarker}{\pgfqpoint{0.000000in}{-0.048611in}}{\pgfqpoint{0.000000in}{0.000000in}}{%
\pgfpathmoveto{\pgfqpoint{0.000000in}{0.000000in}}%
\pgfpathlineto{\pgfqpoint{0.000000in}{-0.048611in}}%
\pgfusepath{stroke,fill}%
}%
\begin{pgfscope}%
\pgfsys@transformshift{1.932432in}{0.749804in}%
\pgfsys@useobject{currentmarker}{}%
\end{pgfscope}%
\end{pgfscope}%
\begin{pgfscope}%
\definecolor{textcolor}{rgb}{0.000000,0.000000,0.000000}%
\pgfsetstrokecolor{textcolor}%
\pgfsetfillcolor{textcolor}%
\pgftext[x=1.932432in,y=0.652581in,,top]{\color{textcolor}\rmfamily\fontsize{18.000000}{21.600000}\selectfont \(\displaystyle {0.2}\)}%
\end{pgfscope}%
\begin{pgfscope}%
\pgfpathrectangle{\pgfqpoint{0.794080in}{0.749804in}}{\pgfqpoint{5.122587in}{5.122587in}}%
\pgfusepath{clip}%
\pgfsetrectcap%
\pgfsetroundjoin%
\pgfsetlinewidth{0.803000pt}%
\definecolor{currentstroke}{rgb}{0.690196,0.690196,0.690196}%
\pgfsetstrokecolor{currentstroke}%
\pgfsetdash{}{0pt}%
\pgfpathmoveto{\pgfqpoint{2.881060in}{0.749804in}}%
\pgfpathlineto{\pgfqpoint{2.881060in}{5.872390in}}%
\pgfusepath{stroke}%
\end{pgfscope}%
\begin{pgfscope}%
\pgfsetbuttcap%
\pgfsetroundjoin%
\definecolor{currentfill}{rgb}{0.000000,0.000000,0.000000}%
\pgfsetfillcolor{currentfill}%
\pgfsetlinewidth{0.803000pt}%
\definecolor{currentstroke}{rgb}{0.000000,0.000000,0.000000}%
\pgfsetstrokecolor{currentstroke}%
\pgfsetdash{}{0pt}%
\pgfsys@defobject{currentmarker}{\pgfqpoint{0.000000in}{-0.048611in}}{\pgfqpoint{0.000000in}{0.000000in}}{%
\pgfpathmoveto{\pgfqpoint{0.000000in}{0.000000in}}%
\pgfpathlineto{\pgfqpoint{0.000000in}{-0.048611in}}%
\pgfusepath{stroke,fill}%
}%
\begin{pgfscope}%
\pgfsys@transformshift{2.881060in}{0.749804in}%
\pgfsys@useobject{currentmarker}{}%
\end{pgfscope}%
\end{pgfscope}%
\begin{pgfscope}%
\definecolor{textcolor}{rgb}{0.000000,0.000000,0.000000}%
\pgfsetstrokecolor{textcolor}%
\pgfsetfillcolor{textcolor}%
\pgftext[x=2.881060in,y=0.652581in,,top]{\color{textcolor}\rmfamily\fontsize{18.000000}{21.600000}\selectfont \(\displaystyle {0.4}\)}%
\end{pgfscope}%
\begin{pgfscope}%
\pgfpathrectangle{\pgfqpoint{0.794080in}{0.749804in}}{\pgfqpoint{5.122587in}{5.122587in}}%
\pgfusepath{clip}%
\pgfsetrectcap%
\pgfsetroundjoin%
\pgfsetlinewidth{0.803000pt}%
\definecolor{currentstroke}{rgb}{0.690196,0.690196,0.690196}%
\pgfsetstrokecolor{currentstroke}%
\pgfsetdash{}{0pt}%
\pgfpathmoveto{\pgfqpoint{3.829687in}{0.749804in}}%
\pgfpathlineto{\pgfqpoint{3.829687in}{5.872390in}}%
\pgfusepath{stroke}%
\end{pgfscope}%
\begin{pgfscope}%
\pgfsetbuttcap%
\pgfsetroundjoin%
\definecolor{currentfill}{rgb}{0.000000,0.000000,0.000000}%
\pgfsetfillcolor{currentfill}%
\pgfsetlinewidth{0.803000pt}%
\definecolor{currentstroke}{rgb}{0.000000,0.000000,0.000000}%
\pgfsetstrokecolor{currentstroke}%
\pgfsetdash{}{0pt}%
\pgfsys@defobject{currentmarker}{\pgfqpoint{0.000000in}{-0.048611in}}{\pgfqpoint{0.000000in}{0.000000in}}{%
\pgfpathmoveto{\pgfqpoint{0.000000in}{0.000000in}}%
\pgfpathlineto{\pgfqpoint{0.000000in}{-0.048611in}}%
\pgfusepath{stroke,fill}%
}%
\begin{pgfscope}%
\pgfsys@transformshift{3.829687in}{0.749804in}%
\pgfsys@useobject{currentmarker}{}%
\end{pgfscope}%
\end{pgfscope}%
\begin{pgfscope}%
\definecolor{textcolor}{rgb}{0.000000,0.000000,0.000000}%
\pgfsetstrokecolor{textcolor}%
\pgfsetfillcolor{textcolor}%
\pgftext[x=3.829687in,y=0.652581in,,top]{\color{textcolor}\rmfamily\fontsize{18.000000}{21.600000}\selectfont \(\displaystyle {0.6}\)}%
\end{pgfscope}%
\begin{pgfscope}%
\pgfpathrectangle{\pgfqpoint{0.794080in}{0.749804in}}{\pgfqpoint{5.122587in}{5.122587in}}%
\pgfusepath{clip}%
\pgfsetrectcap%
\pgfsetroundjoin%
\pgfsetlinewidth{0.803000pt}%
\definecolor{currentstroke}{rgb}{0.690196,0.690196,0.690196}%
\pgfsetstrokecolor{currentstroke}%
\pgfsetdash{}{0pt}%
\pgfpathmoveto{\pgfqpoint{4.778314in}{0.749804in}}%
\pgfpathlineto{\pgfqpoint{4.778314in}{5.872390in}}%
\pgfusepath{stroke}%
\end{pgfscope}%
\begin{pgfscope}%
\pgfsetbuttcap%
\pgfsetroundjoin%
\definecolor{currentfill}{rgb}{0.000000,0.000000,0.000000}%
\pgfsetfillcolor{currentfill}%
\pgfsetlinewidth{0.803000pt}%
\definecolor{currentstroke}{rgb}{0.000000,0.000000,0.000000}%
\pgfsetstrokecolor{currentstroke}%
\pgfsetdash{}{0pt}%
\pgfsys@defobject{currentmarker}{\pgfqpoint{0.000000in}{-0.048611in}}{\pgfqpoint{0.000000in}{0.000000in}}{%
\pgfpathmoveto{\pgfqpoint{0.000000in}{0.000000in}}%
\pgfpathlineto{\pgfqpoint{0.000000in}{-0.048611in}}%
\pgfusepath{stroke,fill}%
}%
\begin{pgfscope}%
\pgfsys@transformshift{4.778314in}{0.749804in}%
\pgfsys@useobject{currentmarker}{}%
\end{pgfscope}%
\end{pgfscope}%
\begin{pgfscope}%
\definecolor{textcolor}{rgb}{0.000000,0.000000,0.000000}%
\pgfsetstrokecolor{textcolor}%
\pgfsetfillcolor{textcolor}%
\pgftext[x=4.778314in,y=0.652581in,,top]{\color{textcolor}\rmfamily\fontsize{18.000000}{21.600000}\selectfont \(\displaystyle {0.8}\)}%
\end{pgfscope}%
\begin{pgfscope}%
\pgfpathrectangle{\pgfqpoint{0.794080in}{0.749804in}}{\pgfqpoint{5.122587in}{5.122587in}}%
\pgfusepath{clip}%
\pgfsetrectcap%
\pgfsetroundjoin%
\pgfsetlinewidth{0.803000pt}%
\definecolor{currentstroke}{rgb}{0.690196,0.690196,0.690196}%
\pgfsetstrokecolor{currentstroke}%
\pgfsetdash{}{0pt}%
\pgfpathmoveto{\pgfqpoint{5.726941in}{0.749804in}}%
\pgfpathlineto{\pgfqpoint{5.726941in}{5.872390in}}%
\pgfusepath{stroke}%
\end{pgfscope}%
\begin{pgfscope}%
\pgfsetbuttcap%
\pgfsetroundjoin%
\definecolor{currentfill}{rgb}{0.000000,0.000000,0.000000}%
\pgfsetfillcolor{currentfill}%
\pgfsetlinewidth{0.803000pt}%
\definecolor{currentstroke}{rgb}{0.000000,0.000000,0.000000}%
\pgfsetstrokecolor{currentstroke}%
\pgfsetdash{}{0pt}%
\pgfsys@defobject{currentmarker}{\pgfqpoint{0.000000in}{-0.048611in}}{\pgfqpoint{0.000000in}{0.000000in}}{%
\pgfpathmoveto{\pgfqpoint{0.000000in}{0.000000in}}%
\pgfpathlineto{\pgfqpoint{0.000000in}{-0.048611in}}%
\pgfusepath{stroke,fill}%
}%
\begin{pgfscope}%
\pgfsys@transformshift{5.726941in}{0.749804in}%
\pgfsys@useobject{currentmarker}{}%
\end{pgfscope}%
\end{pgfscope}%
\begin{pgfscope}%
\definecolor{textcolor}{rgb}{0.000000,0.000000,0.000000}%
\pgfsetstrokecolor{textcolor}%
\pgfsetfillcolor{textcolor}%
\pgftext[x=5.726941in,y=0.652581in,,top]{\color{textcolor}\rmfamily\fontsize{18.000000}{21.600000}\selectfont \(\displaystyle {1.0}\)}%
\end{pgfscope}%
\begin{pgfscope}%
\definecolor{textcolor}{rgb}{0.000000,0.000000,0.000000}%
\pgfsetstrokecolor{textcolor}%
\pgfsetfillcolor{textcolor}%
\pgftext[x=3.355373in,y=0.383677in,,top]{\color{textcolor}\rmfamily\fontsize{20.000000}{24.000000}\selectfont Confidence level \(\displaystyle 1 - \alpha\)}%
\end{pgfscope}%
\begin{pgfscope}%
\pgfpathrectangle{\pgfqpoint{0.794080in}{0.749804in}}{\pgfqpoint{5.122587in}{5.122587in}}%
\pgfusepath{clip}%
\pgfsetrectcap%
\pgfsetroundjoin%
\pgfsetlinewidth{0.803000pt}%
\definecolor{currentstroke}{rgb}{0.690196,0.690196,0.690196}%
\pgfsetstrokecolor{currentstroke}%
\pgfsetdash{}{0pt}%
\pgfpathmoveto{\pgfqpoint{0.794080in}{0.939529in}}%
\pgfpathlineto{\pgfqpoint{5.916667in}{0.939529in}}%
\pgfusepath{stroke}%
\end{pgfscope}%
\begin{pgfscope}%
\pgfsetbuttcap%
\pgfsetroundjoin%
\definecolor{currentfill}{rgb}{0.000000,0.000000,0.000000}%
\pgfsetfillcolor{currentfill}%
\pgfsetlinewidth{0.803000pt}%
\definecolor{currentstroke}{rgb}{0.000000,0.000000,0.000000}%
\pgfsetstrokecolor{currentstroke}%
\pgfsetdash{}{0pt}%
\pgfsys@defobject{currentmarker}{\pgfqpoint{-0.048611in}{0.000000in}}{\pgfqpoint{-0.000000in}{0.000000in}}{%
\pgfpathmoveto{\pgfqpoint{-0.000000in}{0.000000in}}%
\pgfpathlineto{\pgfqpoint{-0.048611in}{0.000000in}}%
\pgfusepath{stroke,fill}%
}%
\begin{pgfscope}%
\pgfsys@transformshift{0.794080in}{0.939529in}%
\pgfsys@useobject{currentmarker}{}%
\end{pgfscope}%
\end{pgfscope}%
\begin{pgfscope}%
\definecolor{textcolor}{rgb}{0.000000,0.000000,0.000000}%
\pgfsetstrokecolor{textcolor}%
\pgfsetfillcolor{textcolor}%
\pgftext[x=0.411444in, y=0.856196in, left, base]{\color{textcolor}\rmfamily\fontsize{18.000000}{21.600000}\selectfont \(\displaystyle {0.0}\)}%
\end{pgfscope}%
\begin{pgfscope}%
\pgfpathrectangle{\pgfqpoint{0.794080in}{0.749804in}}{\pgfqpoint{5.122587in}{5.122587in}}%
\pgfusepath{clip}%
\pgfsetrectcap%
\pgfsetroundjoin%
\pgfsetlinewidth{0.803000pt}%
\definecolor{currentstroke}{rgb}{0.690196,0.690196,0.690196}%
\pgfsetstrokecolor{currentstroke}%
\pgfsetdash{}{0pt}%
\pgfpathmoveto{\pgfqpoint{0.794080in}{1.888156in}}%
\pgfpathlineto{\pgfqpoint{5.916667in}{1.888156in}}%
\pgfusepath{stroke}%
\end{pgfscope}%
\begin{pgfscope}%
\pgfsetbuttcap%
\pgfsetroundjoin%
\definecolor{currentfill}{rgb}{0.000000,0.000000,0.000000}%
\pgfsetfillcolor{currentfill}%
\pgfsetlinewidth{0.803000pt}%
\definecolor{currentstroke}{rgb}{0.000000,0.000000,0.000000}%
\pgfsetstrokecolor{currentstroke}%
\pgfsetdash{}{0pt}%
\pgfsys@defobject{currentmarker}{\pgfqpoint{-0.048611in}{0.000000in}}{\pgfqpoint{-0.000000in}{0.000000in}}{%
\pgfpathmoveto{\pgfqpoint{-0.000000in}{0.000000in}}%
\pgfpathlineto{\pgfqpoint{-0.048611in}{0.000000in}}%
\pgfusepath{stroke,fill}%
}%
\begin{pgfscope}%
\pgfsys@transformshift{0.794080in}{1.888156in}%
\pgfsys@useobject{currentmarker}{}%
\end{pgfscope}%
\end{pgfscope}%
\begin{pgfscope}%
\definecolor{textcolor}{rgb}{0.000000,0.000000,0.000000}%
\pgfsetstrokecolor{textcolor}%
\pgfsetfillcolor{textcolor}%
\pgftext[x=0.411444in, y=1.804823in, left, base]{\color{textcolor}\rmfamily\fontsize{18.000000}{21.600000}\selectfont \(\displaystyle {0.2}\)}%
\end{pgfscope}%
\begin{pgfscope}%
\pgfpathrectangle{\pgfqpoint{0.794080in}{0.749804in}}{\pgfqpoint{5.122587in}{5.122587in}}%
\pgfusepath{clip}%
\pgfsetrectcap%
\pgfsetroundjoin%
\pgfsetlinewidth{0.803000pt}%
\definecolor{currentstroke}{rgb}{0.690196,0.690196,0.690196}%
\pgfsetstrokecolor{currentstroke}%
\pgfsetdash{}{0pt}%
\pgfpathmoveto{\pgfqpoint{0.794080in}{2.836783in}}%
\pgfpathlineto{\pgfqpoint{5.916667in}{2.836783in}}%
\pgfusepath{stroke}%
\end{pgfscope}%
\begin{pgfscope}%
\pgfsetbuttcap%
\pgfsetroundjoin%
\definecolor{currentfill}{rgb}{0.000000,0.000000,0.000000}%
\pgfsetfillcolor{currentfill}%
\pgfsetlinewidth{0.803000pt}%
\definecolor{currentstroke}{rgb}{0.000000,0.000000,0.000000}%
\pgfsetstrokecolor{currentstroke}%
\pgfsetdash{}{0pt}%
\pgfsys@defobject{currentmarker}{\pgfqpoint{-0.048611in}{0.000000in}}{\pgfqpoint{-0.000000in}{0.000000in}}{%
\pgfpathmoveto{\pgfqpoint{-0.000000in}{0.000000in}}%
\pgfpathlineto{\pgfqpoint{-0.048611in}{0.000000in}}%
\pgfusepath{stroke,fill}%
}%
\begin{pgfscope}%
\pgfsys@transformshift{0.794080in}{2.836783in}%
\pgfsys@useobject{currentmarker}{}%
\end{pgfscope}%
\end{pgfscope}%
\begin{pgfscope}%
\definecolor{textcolor}{rgb}{0.000000,0.000000,0.000000}%
\pgfsetstrokecolor{textcolor}%
\pgfsetfillcolor{textcolor}%
\pgftext[x=0.411444in, y=2.753450in, left, base]{\color{textcolor}\rmfamily\fontsize{18.000000}{21.600000}\selectfont \(\displaystyle {0.4}\)}%
\end{pgfscope}%
\begin{pgfscope}%
\pgfpathrectangle{\pgfqpoint{0.794080in}{0.749804in}}{\pgfqpoint{5.122587in}{5.122587in}}%
\pgfusepath{clip}%
\pgfsetrectcap%
\pgfsetroundjoin%
\pgfsetlinewidth{0.803000pt}%
\definecolor{currentstroke}{rgb}{0.690196,0.690196,0.690196}%
\pgfsetstrokecolor{currentstroke}%
\pgfsetdash{}{0pt}%
\pgfpathmoveto{\pgfqpoint{0.794080in}{3.785411in}}%
\pgfpathlineto{\pgfqpoint{5.916667in}{3.785411in}}%
\pgfusepath{stroke}%
\end{pgfscope}%
\begin{pgfscope}%
\pgfsetbuttcap%
\pgfsetroundjoin%
\definecolor{currentfill}{rgb}{0.000000,0.000000,0.000000}%
\pgfsetfillcolor{currentfill}%
\pgfsetlinewidth{0.803000pt}%
\definecolor{currentstroke}{rgb}{0.000000,0.000000,0.000000}%
\pgfsetstrokecolor{currentstroke}%
\pgfsetdash{}{0pt}%
\pgfsys@defobject{currentmarker}{\pgfqpoint{-0.048611in}{0.000000in}}{\pgfqpoint{-0.000000in}{0.000000in}}{%
\pgfpathmoveto{\pgfqpoint{-0.000000in}{0.000000in}}%
\pgfpathlineto{\pgfqpoint{-0.048611in}{0.000000in}}%
\pgfusepath{stroke,fill}%
}%
\begin{pgfscope}%
\pgfsys@transformshift{0.794080in}{3.785411in}%
\pgfsys@useobject{currentmarker}{}%
\end{pgfscope}%
\end{pgfscope}%
\begin{pgfscope}%
\definecolor{textcolor}{rgb}{0.000000,0.000000,0.000000}%
\pgfsetstrokecolor{textcolor}%
\pgfsetfillcolor{textcolor}%
\pgftext[x=0.411444in, y=3.702077in, left, base]{\color{textcolor}\rmfamily\fontsize{18.000000}{21.600000}\selectfont \(\displaystyle {0.6}\)}%
\end{pgfscope}%
\begin{pgfscope}%
\pgfpathrectangle{\pgfqpoint{0.794080in}{0.749804in}}{\pgfqpoint{5.122587in}{5.122587in}}%
\pgfusepath{clip}%
\pgfsetrectcap%
\pgfsetroundjoin%
\pgfsetlinewidth{0.803000pt}%
\definecolor{currentstroke}{rgb}{0.690196,0.690196,0.690196}%
\pgfsetstrokecolor{currentstroke}%
\pgfsetdash{}{0pt}%
\pgfpathmoveto{\pgfqpoint{0.794080in}{4.734038in}}%
\pgfpathlineto{\pgfqpoint{5.916667in}{4.734038in}}%
\pgfusepath{stroke}%
\end{pgfscope}%
\begin{pgfscope}%
\pgfsetbuttcap%
\pgfsetroundjoin%
\definecolor{currentfill}{rgb}{0.000000,0.000000,0.000000}%
\pgfsetfillcolor{currentfill}%
\pgfsetlinewidth{0.803000pt}%
\definecolor{currentstroke}{rgb}{0.000000,0.000000,0.000000}%
\pgfsetstrokecolor{currentstroke}%
\pgfsetdash{}{0pt}%
\pgfsys@defobject{currentmarker}{\pgfqpoint{-0.048611in}{0.000000in}}{\pgfqpoint{-0.000000in}{0.000000in}}{%
\pgfpathmoveto{\pgfqpoint{-0.000000in}{0.000000in}}%
\pgfpathlineto{\pgfqpoint{-0.048611in}{0.000000in}}%
\pgfusepath{stroke,fill}%
}%
\begin{pgfscope}%
\pgfsys@transformshift{0.794080in}{4.734038in}%
\pgfsys@useobject{currentmarker}{}%
\end{pgfscope}%
\end{pgfscope}%
\begin{pgfscope}%
\definecolor{textcolor}{rgb}{0.000000,0.000000,0.000000}%
\pgfsetstrokecolor{textcolor}%
\pgfsetfillcolor{textcolor}%
\pgftext[x=0.411444in, y=4.650704in, left, base]{\color{textcolor}\rmfamily\fontsize{18.000000}{21.600000}\selectfont \(\displaystyle {0.8}\)}%
\end{pgfscope}%
\begin{pgfscope}%
\pgfpathrectangle{\pgfqpoint{0.794080in}{0.749804in}}{\pgfqpoint{5.122587in}{5.122587in}}%
\pgfusepath{clip}%
\pgfsetrectcap%
\pgfsetroundjoin%
\pgfsetlinewidth{0.803000pt}%
\definecolor{currentstroke}{rgb}{0.690196,0.690196,0.690196}%
\pgfsetstrokecolor{currentstroke}%
\pgfsetdash{}{0pt}%
\pgfpathmoveto{\pgfqpoint{0.794080in}{5.682665in}}%
\pgfpathlineto{\pgfqpoint{5.916667in}{5.682665in}}%
\pgfusepath{stroke}%
\end{pgfscope}%
\begin{pgfscope}%
\pgfsetbuttcap%
\pgfsetroundjoin%
\definecolor{currentfill}{rgb}{0.000000,0.000000,0.000000}%
\pgfsetfillcolor{currentfill}%
\pgfsetlinewidth{0.803000pt}%
\definecolor{currentstroke}{rgb}{0.000000,0.000000,0.000000}%
\pgfsetstrokecolor{currentstroke}%
\pgfsetdash{}{0pt}%
\pgfsys@defobject{currentmarker}{\pgfqpoint{-0.048611in}{0.000000in}}{\pgfqpoint{-0.000000in}{0.000000in}}{%
\pgfpathmoveto{\pgfqpoint{-0.000000in}{0.000000in}}%
\pgfpathlineto{\pgfqpoint{-0.048611in}{0.000000in}}%
\pgfusepath{stroke,fill}%
}%
\begin{pgfscope}%
\pgfsys@transformshift{0.794080in}{5.682665in}%
\pgfsys@useobject{currentmarker}{}%
\end{pgfscope}%
\end{pgfscope}%
\begin{pgfscope}%
\definecolor{textcolor}{rgb}{0.000000,0.000000,0.000000}%
\pgfsetstrokecolor{textcolor}%
\pgfsetfillcolor{textcolor}%
\pgftext[x=0.411444in, y=5.599332in, left, base]{\color{textcolor}\rmfamily\fontsize{18.000000}{21.600000}\selectfont \(\displaystyle {1.0}\)}%
\end{pgfscope}%
\begin{pgfscope}%
\definecolor{textcolor}{rgb}{0.000000,0.000000,0.000000}%
\pgfsetstrokecolor{textcolor}%
\pgfsetfillcolor{textcolor}%
\pgftext[x=0.355889in,y=3.311097in,,bottom,rotate=90.000000]{\color{textcolor}\rmfamily\fontsize{20.000000}{24.000000}\selectfont \(\displaystyle \mathbb{P}\)(accept \(\displaystyle H_0\))}%
\end{pgfscope}%
\begin{pgfscope}%
\pgfpathrectangle{\pgfqpoint{0.794080in}{0.749804in}}{\pgfqpoint{5.122587in}{5.122587in}}%
\pgfusepath{clip}%
\pgfsetbuttcap%
\pgfsetroundjoin%
\pgfsetlinewidth{1.505625pt}%
\definecolor{currentstroke}{rgb}{0.411765,0.411765,0.411765}%
\pgfsetstrokecolor{currentstroke}%
\pgfsetdash{{1.500000pt}{2.475000pt}}{0.000000pt}%
\pgfpathmoveto{\pgfqpoint{5.489784in}{0.739804in}}%
\pgfpathlineto{\pgfqpoint{5.489784in}{5.882390in}}%
\pgfusepath{stroke}%
\end{pgfscope}%
\begin{pgfscope}%
\pgfpathrectangle{\pgfqpoint{0.794080in}{0.749804in}}{\pgfqpoint{5.122587in}{5.122587in}}%
\pgfusepath{clip}%
\pgfsetrectcap%
\pgfsetroundjoin%
\pgfsetlinewidth{1.505625pt}%
\definecolor{currentstroke}{rgb}{0.000000,0.000000,1.000000}%
\pgfsetstrokecolor{currentstroke}%
\pgfsetdash{}{0pt}%
\pgfpathmoveto{\pgfqpoint{0.983805in}{0.939529in}}%
\pgfpathlineto{\pgfqpoint{1.050209in}{0.969247in}}%
\pgfpathlineto{\pgfqpoint{1.135586in}{1.007503in}}%
\pgfpathlineto{\pgfqpoint{1.220962in}{1.044680in}}%
\pgfpathlineto{\pgfqpoint{1.273136in}{1.067584in}}%
\pgfpathlineto{\pgfqpoint{1.529266in}{1.180163in}}%
\pgfpathlineto{\pgfqpoint{1.557725in}{1.192967in}}%
\pgfpathlineto{\pgfqpoint{1.619385in}{1.220633in}}%
\pgfpathlineto{\pgfqpoint{1.756936in}{1.281998in}}%
\pgfpathlineto{\pgfqpoint{1.837570in}{1.318335in}}%
\pgfpathlineto{\pgfqpoint{2.169589in}{1.467774in}}%
\pgfpathlineto{\pgfqpoint{2.245479in}{1.502475in}}%
\pgfpathlineto{\pgfqpoint{2.364058in}{1.556813in}}%
\pgfpathlineto{\pgfqpoint{2.610701in}{1.672280in}}%
\pgfpathlineto{\pgfqpoint{2.662875in}{1.697058in}}%
\pgfpathlineto{\pgfqpoint{2.805169in}{1.765077in}}%
\pgfpathlineto{\pgfqpoint{2.990152in}{1.855220in}}%
\pgfpathlineto{\pgfqpoint{3.047069in}{1.883321in}}%
\pgfpathlineto{\pgfqpoint{3.269997in}{1.995356in}}%
\pgfpathlineto{\pgfqpoint{3.355373in}{2.039431in}}%
\pgfpathlineto{\pgfqpoint{3.440750in}{2.084157in}}%
\pgfpathlineto{\pgfqpoint{3.483438in}{2.106784in}}%
\pgfpathlineto{\pgfqpoint{3.530869in}{2.131721in}}%
\pgfpathlineto{\pgfqpoint{3.725338in}{2.238223in}}%
\pgfpathlineto{\pgfqpoint{3.976724in}{2.382035in}}%
\pgfpathlineto{\pgfqpoint{4.171193in}{2.499890in}}%
\pgfpathlineto{\pgfqpoint{4.275542in}{2.566515in}}%
\pgfpathlineto{\pgfqpoint{4.436808in}{2.673980in}}%
\pgfpathlineto{\pgfqpoint{4.588589in}{2.781896in}}%
\pgfpathlineto{\pgfqpoint{4.669222in}{2.842506in}}%
\pgfpathlineto{\pgfqpoint{4.740369in}{2.897794in}}%
\pgfpathlineto{\pgfqpoint{4.863690in}{2.999029in}}%
\pgfpathlineto{\pgfqpoint{4.991755in}{3.112969in}}%
\pgfpathlineto{\pgfqpoint{5.053416in}{3.171867in}}%
\pgfpathlineto{\pgfqpoint{5.096104in}{3.214721in}}%
\pgfpathlineto{\pgfqpoint{5.138792in}{3.258991in}}%
\pgfpathlineto{\pgfqpoint{5.195710in}{3.320844in}}%
\pgfpathlineto{\pgfqpoint{5.238398in}{3.370296in}}%
\pgfpathlineto{\pgfqpoint{5.276343in}{3.416968in}}%
\pgfpathlineto{\pgfqpoint{5.309545in}{3.459450in}}%
\pgfpathlineto{\pgfqpoint{5.361720in}{3.531143in}}%
\pgfpathlineto{\pgfqpoint{5.409151in}{3.601868in}}%
\pgfpathlineto{\pgfqpoint{5.456582in}{3.679411in}}%
\pgfpathlineto{\pgfqpoint{5.480298in}{3.721146in}}%
\pgfpathlineto{\pgfqpoint{5.508757in}{3.775417in}}%
\pgfpathlineto{\pgfqpoint{5.537216in}{3.834841in}}%
\pgfpathlineto{\pgfqpoint{5.560931in}{3.888685in}}%
\pgfpathlineto{\pgfqpoint{5.589390in}{3.960923in}}%
\pgfpathlineto{\pgfqpoint{5.613106in}{4.029065in}}%
\pgfpathlineto{\pgfqpoint{5.632079in}{4.091846in}}%
\pgfpathlineto{\pgfqpoint{5.651051in}{4.162903in}}%
\pgfpathlineto{\pgfqpoint{5.665280in}{4.224748in}}%
\pgfpathlineto{\pgfqpoint{5.679510in}{4.297601in}}%
\pgfpathlineto{\pgfqpoint{5.688996in}{4.355650in}}%
\pgfpathlineto{\pgfqpoint{5.698482in}{4.425004in}}%
\pgfpathlineto{\pgfqpoint{5.707969in}{4.512941in}}%
\pgfpathlineto{\pgfqpoint{5.712712in}{4.570205in}}%
\pgfpathlineto{\pgfqpoint{5.717455in}{4.643046in}}%
\pgfpathlineto{\pgfqpoint{5.722198in}{4.750406in}}%
\pgfpathlineto{\pgfqpoint{5.726941in}{5.682665in}}%
\pgfpathlineto{\pgfqpoint{5.726941in}{5.682665in}}%
\pgfusepath{stroke}%
\end{pgfscope}%
\begin{pgfscope}%
\pgfpathrectangle{\pgfqpoint{0.794080in}{0.749804in}}{\pgfqpoint{5.122587in}{5.122587in}}%
\pgfusepath{clip}%
\pgfsetrectcap%
\pgfsetroundjoin%
\pgfsetlinewidth{1.505625pt}%
\definecolor{currentstroke}{rgb}{1.000000,0.300000,0.300000}%
\pgfsetstrokecolor{currentstroke}%
\pgfsetdash{}{0pt}%
\pgfpathmoveto{\pgfqpoint{0.983805in}{0.939529in}}%
\pgfpathlineto{\pgfqpoint{1.040723in}{0.997912in}}%
\pgfpathlineto{\pgfqpoint{1.111870in}{1.070202in}}%
\pgfpathlineto{\pgfqpoint{1.372742in}{1.334454in}}%
\pgfpathlineto{\pgfqpoint{1.415431in}{1.377738in}}%
\pgfpathlineto{\pgfqpoint{2.069983in}{2.043280in}}%
\pgfpathlineto{\pgfqpoint{2.131644in}{2.105634in}}%
\pgfpathlineto{\pgfqpoint{2.212277in}{2.187763in}}%
\pgfpathlineto{\pgfqpoint{2.250222in}{2.225668in}}%
\pgfpathlineto{\pgfqpoint{2.297654in}{2.273275in}}%
\pgfpathlineto{\pgfqpoint{2.667618in}{2.646313in}}%
\pgfpathlineto{\pgfqpoint{2.696077in}{2.674779in}}%
\pgfpathlineto{\pgfqpoint{2.795683in}{2.774247in}}%
\pgfpathlineto{\pgfqpoint{2.994895in}{2.974851in}}%
\pgfpathlineto{\pgfqpoint{3.061299in}{3.042014in}}%
\pgfpathlineto{\pgfqpoint{3.137189in}{3.118495in}}%
\pgfpathlineto{\pgfqpoint{3.222565in}{3.204263in}}%
\pgfpathlineto{\pgfqpoint{4.194908in}{4.170802in}}%
\pgfpathlineto{\pgfqpoint{4.247083in}{4.222182in}}%
\pgfpathlineto{\pgfqpoint{4.896892in}{4.855512in}}%
\pgfpathlineto{\pgfqpoint{4.939581in}{4.896832in}}%
\pgfpathlineto{\pgfqpoint{4.987012in}{4.943156in}}%
\pgfpathlineto{\pgfqpoint{5.039186in}{4.994211in}}%
\pgfpathlineto{\pgfqpoint{5.252628in}{5.199325in}}%
\pgfpathlineto{\pgfqpoint{5.285830in}{5.231398in}}%
\pgfpathlineto{\pgfqpoint{5.537216in}{5.473286in}}%
\pgfpathlineto{\pgfqpoint{5.613106in}{5.546483in}}%
\pgfpathlineto{\pgfqpoint{5.679510in}{5.612716in}}%
\pgfpathlineto{\pgfqpoint{5.717455in}{5.655693in}}%
\pgfpathlineto{\pgfqpoint{5.722198in}{5.662379in}}%
\pgfpathlineto{\pgfqpoint{5.726941in}{5.682665in}}%
\pgfpathlineto{\pgfqpoint{5.726941in}{5.682665in}}%
\pgfusepath{stroke}%
\end{pgfscope}%
\begin{pgfscope}%
\pgfpathrectangle{\pgfqpoint{0.794080in}{0.749804in}}{\pgfqpoint{5.122587in}{5.122587in}}%
\pgfusepath{clip}%
\pgfsetbuttcap%
\pgfsetroundjoin%
\pgfsetlinewidth{1.505625pt}%
\definecolor{currentstroke}{rgb}{0.000000,0.000000,0.000000}%
\pgfsetstrokecolor{currentstroke}%
\pgfsetdash{{7.500000pt}{7.500000pt}}{0.000000pt}%
\pgfpathmoveto{\pgfqpoint{0.983805in}{0.939529in}}%
\pgfpathlineto{\pgfqpoint{5.726941in}{5.682665in}}%
\pgfpathlineto{\pgfqpoint{5.726941in}{5.682665in}}%
\pgfusepath{stroke}%
\end{pgfscope}%
\begin{pgfscope}%
\pgfsetrectcap%
\pgfsetmiterjoin%
\pgfsetlinewidth{0.803000pt}%
\definecolor{currentstroke}{rgb}{0.000000,0.000000,0.000000}%
\pgfsetstrokecolor{currentstroke}%
\pgfsetdash{}{0pt}%
\pgfpathmoveto{\pgfqpoint{0.794080in}{0.749804in}}%
\pgfpathlineto{\pgfqpoint{0.794080in}{5.872390in}}%
\pgfusepath{stroke}%
\end{pgfscope}%
\begin{pgfscope}%
\pgfsetrectcap%
\pgfsetmiterjoin%
\pgfsetlinewidth{0.803000pt}%
\definecolor{currentstroke}{rgb}{0.000000,0.000000,0.000000}%
\pgfsetstrokecolor{currentstroke}%
\pgfsetdash{}{0pt}%
\pgfpathmoveto{\pgfqpoint{5.916667in}{0.749804in}}%
\pgfpathlineto{\pgfqpoint{5.916667in}{5.872390in}}%
\pgfusepath{stroke}%
\end{pgfscope}%
\begin{pgfscope}%
\pgfsetrectcap%
\pgfsetmiterjoin%
\pgfsetlinewidth{0.803000pt}%
\definecolor{currentstroke}{rgb}{0.000000,0.000000,0.000000}%
\pgfsetstrokecolor{currentstroke}%
\pgfsetdash{}{0pt}%
\pgfpathmoveto{\pgfqpoint{0.794080in}{0.749804in}}%
\pgfpathlineto{\pgfqpoint{5.916667in}{0.749804in}}%
\pgfusepath{stroke}%
\end{pgfscope}%
\begin{pgfscope}%
\pgfsetrectcap%
\pgfsetmiterjoin%
\pgfsetlinewidth{0.803000pt}%
\definecolor{currentstroke}{rgb}{0.000000,0.000000,0.000000}%
\pgfsetstrokecolor{currentstroke}%
\pgfsetdash{}{0pt}%
\pgfpathmoveto{\pgfqpoint{0.794080in}{5.872390in}}%
\pgfpathlineto{\pgfqpoint{5.916667in}{5.872390in}}%
\pgfusepath{stroke}%
\end{pgfscope}%
\begin{pgfscope}%
\definecolor{textcolor}{rgb}{0.411765,0.411765,0.411765}%
\pgfsetstrokecolor{textcolor}%
\pgfsetfillcolor{textcolor}%
\pgftext[x=4.682081in, y=1.472761in, left, base]{\color{textcolor}\rmfamily\fontsize{20.000000}{24.000000}\selectfont \(\displaystyle 1 - \alpha\)}%
\end{pgfscope}%
\begin{pgfscope}%
\definecolor{textcolor}{rgb}{0.411765,0.411765,0.411765}%
\pgfsetstrokecolor{textcolor}%
\pgfsetfillcolor{textcolor}%
\pgftext[x=4.573505in, y=1.176686in, left, base]{\color{textcolor}\rmfamily\fontsize{20.000000}{24.000000}\selectfont  \(\displaystyle = 95\%\)}%
\end{pgfscope}%
\begin{pgfscope}%
\pgfsetbuttcap%
\pgfsetmiterjoin%
\definecolor{currentfill}{rgb}{1.000000,1.000000,1.000000}%
\pgfsetfillcolor{currentfill}%
\pgfsetfillopacity{0.800000}%
\pgfsetlinewidth{1.003750pt}%
\definecolor{currentstroke}{rgb}{0.800000,0.800000,0.800000}%
\pgfsetstrokecolor{currentstroke}%
\pgfsetstrokeopacity{0.800000}%
\pgfsetdash{}{0pt}%
\pgfpathmoveto{\pgfqpoint{0.988524in}{4.828768in}}%
\pgfpathlineto{\pgfqpoint{3.954298in}{4.828768in}}%
\pgfpathquadraticcurveto{\pgfqpoint{4.009854in}{4.828768in}}{\pgfqpoint{4.009854in}{4.884323in}}%
\pgfpathlineto{\pgfqpoint{4.009854in}{5.677946in}}%
\pgfpathquadraticcurveto{\pgfqpoint{4.009854in}{5.733502in}}{\pgfqpoint{3.954298in}{5.733502in}}%
\pgfpathlineto{\pgfqpoint{0.988524in}{5.733502in}}%
\pgfpathquadraticcurveto{\pgfqpoint{0.932969in}{5.733502in}}{\pgfqpoint{0.932969in}{5.677946in}}%
\pgfpathlineto{\pgfqpoint{0.932969in}{4.884323in}}%
\pgfpathquadraticcurveto{\pgfqpoint{0.932969in}{4.828768in}}{\pgfqpoint{0.988524in}{4.828768in}}%
\pgfpathclose%
\pgfusepath{stroke,fill}%
\end{pgfscope}%
\begin{pgfscope}%
\pgfsetrectcap%
\pgfsetroundjoin%
\pgfsetlinewidth{1.505625pt}%
\definecolor{currentstroke}{rgb}{1.000000,0.300000,0.300000}%
\pgfsetstrokecolor{currentstroke}%
\pgfsetdash{}{0pt}%
\pgfpathmoveto{\pgfqpoint{1.044080in}{5.503830in}}%
\pgfpathlineto{\pgfqpoint{1.599635in}{5.503830in}}%
\pgfusepath{stroke}%
\end{pgfscope}%
\begin{pgfscope}%
\definecolor{textcolor}{rgb}{0.000000,0.000000,0.000000}%
\pgfsetstrokecolor{textcolor}%
\pgfsetfillcolor{textcolor}%
\pgftext[x=1.821857in,y=5.406608in,left,base]{\color{textcolor}\rmfamily\fontsize{20.000000}{24.000000}\selectfont NEBP (proposed)}%
\end{pgfscope}%
\begin{pgfscope}%
\pgfsetrectcap%
\pgfsetroundjoin%
\pgfsetlinewidth{1.505625pt}%
\definecolor{currentstroke}{rgb}{0.000000,0.000000,1.000000}%
\pgfsetstrokecolor{currentstroke}%
\pgfsetdash{}{0pt}%
\pgfpathmoveto{\pgfqpoint{1.044080in}{5.093130in}}%
\pgfpathlineto{\pgfqpoint{1.599635in}{5.093130in}}%
\pgfusepath{stroke}%
\end{pgfscope}%
\begin{pgfscope}%
\definecolor{textcolor}{rgb}{0.000000,0.000000,0.000000}%
\pgfsetstrokecolor{textcolor}%
\pgfsetfillcolor{textcolor}%
\pgftext[x=1.821857in,y=4.995908in,left,base]{\color{textcolor}\rmfamily\fontsize{20.000000}{24.000000}\selectfont BP}%
\end{pgfscope}%
\end{pgfpicture}%
\makeatother%
\endgroup%